\documentclass[10pt,twocolumn,letterpaper]{article}
 
\usepackage[pagenumbers]{wacv}

\makeatletter
\@namedef{ver@everyshi.sty}{}
\makeatother

\usepackage[accsupp]{axessibility}  %
\usepackage{graphicx}
\usepackage{amsmath}
\usepackage{amssymb}
\usepackage{booktabs}
\usepackage{gensymb}
\usepackage[acronym]{glossaries}
\usepackage{multirow, hhline}
\usepackage[symbol]{footmisc}
\usepackage[textsize=footnotesize]{todonotes}
\usepackage{xcolor,colortbl}

\usepackage{caption}
\usepackage{subcaption}
\usepackage{forloop}

\makeatletter
\newcommand{\twodigits}{\two@digits}
\makeatother
\newcolumntype{P}[1]{>{\centering\arraybackslash}p{#1}}
\newcolumntype{M}[1]{>{\centering\arraybackslash}m{#1}}
\newcolumntype{x}[1]{>{\centering\let\newline\\\arraybackslash\hspace{0pt}}p{#1}}
\newcolumntype{H}{>{\setbox0=\hbox\bgroup}c<{\egroup}@{}}

\usepackage{pgfplots}
\usepgfplotslibrary{fillbetween}

\makeglossaries
\newacronym{bev}{BEV}{bird's-eye view}
\newacronym[plural=SSC,firstplural=submanifold sparse convolutions (SSC)]{ssc}{SSC}{submanifold sparse convolution}
\newacronym[plural=SC,firstplural=sparse convolutions (SC)]{sc}{SC}{sparse convolution}
\newacronym{mae}{MAE}{masked autoencoder}
\newacronym{ssrl}{SSRL}{self-supervised representation learning}
\newacronym{nlp}{NLP}{natural language processing}
\newacronym{fps}{FPS}{Farthest Point Sampling}
\newacronym{knn}{KNN}{K-Nearest Neighbor}
\newacronym{wod}{WOD}{Waymo Open Dataset}

\usepackage[pagebackref,breaklinks,colorlinks]{hyperref}

\usepackage[capitalize]{cleveref}
\crefname{section}{Sec.}{Secs.}
\Crefname{section}{Section}{Sections}
\Crefname{table}{Table}{Tables}
\crefname{table}{Tab.}{Tabs.}

\definecolor{petrol}{HTML}{39979c}

\def\confName{WACV}
\def\confYear{2024}

\begin{document}

\title{MAELi: Masked Autoencoder for Large-Scale LiDAR Point Clouds}

\author{
	{Georg Krispel\textsuperscript{1} \hspace{0.0001em}
	David Schinagl\textsuperscript{1,2}  \hspace{0.0001em}
	Christian Fruhwirth-Reisinger\textsuperscript{1,2} \hspace{0.0001em}
	Horst Possegger\textsuperscript{1} \hspace{0.0001em}
	Horst Bischof\textsuperscript{1,2}} \\ 
	{\tt\small\{georg.krispel,david.schinagl,christian.reisinger,possegger,bischof\}@icg.tugraz.at}\\
	\textsuperscript{1} Graz University of Technology\hspace{0.0001em} 
	\textsuperscript{2} Christian Doppler Laboratory for Embedded Machine Learning  \\
}

\maketitle

\begin{abstract}
    The sensing process of large-scale LiDAR point clouds inevitably causes large blind spots, \ie regions not visible to the sensor.
    We demonstrate how these inherent sampling properties can be effectively utilized for self-supervised representation learning by designing
    a highly effective pre-training framework that considerably reduces the need for tedious 3D annotations to train state-of-the-art object detectors.
    Our \emph{Masked AutoEncoder for LiDAR point clouds (MAELi)} intuitively leverages the sparsity of LiDAR point clouds in both the encoder and decoder during reconstruction. 
    This results in more expressive and useful initialization, which can be directly applied to downstream perception tasks, such as 3D object detection or semantic segmentation for autonomous driving.
    In a novel reconstruction approach, MAELi distinguishes between empty and occluded space and employs a new masking strategy that targets the LiDAR's inherent spherical projection. 
    Thereby, without any ground truth whatsoever and trained on single frames only, MAELi obtains an understanding of the underlying 3D scene geometry and semantics.
    To demonstrate the potential of MAELi, we pre-train backbones in an end-to-end manner and show the effectiveness of our unsupervised pre-trained weights on the tasks of 3D object detection and semantic segmentation.
\end{abstract}

\section{Introduction}
\label{sec:intro}

Thanks to recent large-scale and elaborately curated datasets, such as the Waymo Open Dataset~\cite{sunScalabilityPerceptionAutonomous2020}, we have witnessed tremendous progress in a diverse variety of 3D perception tasks crucial for autonomous driving. 
However, even with the aid of such cost-intensive datasets, models remain only transferable to other domains while suffering significant performance drops~\cite{wangTrainGermanyTest2020}.

\begin{figure}[h!]
	\centering
 	\includegraphics[width=\linewidth]{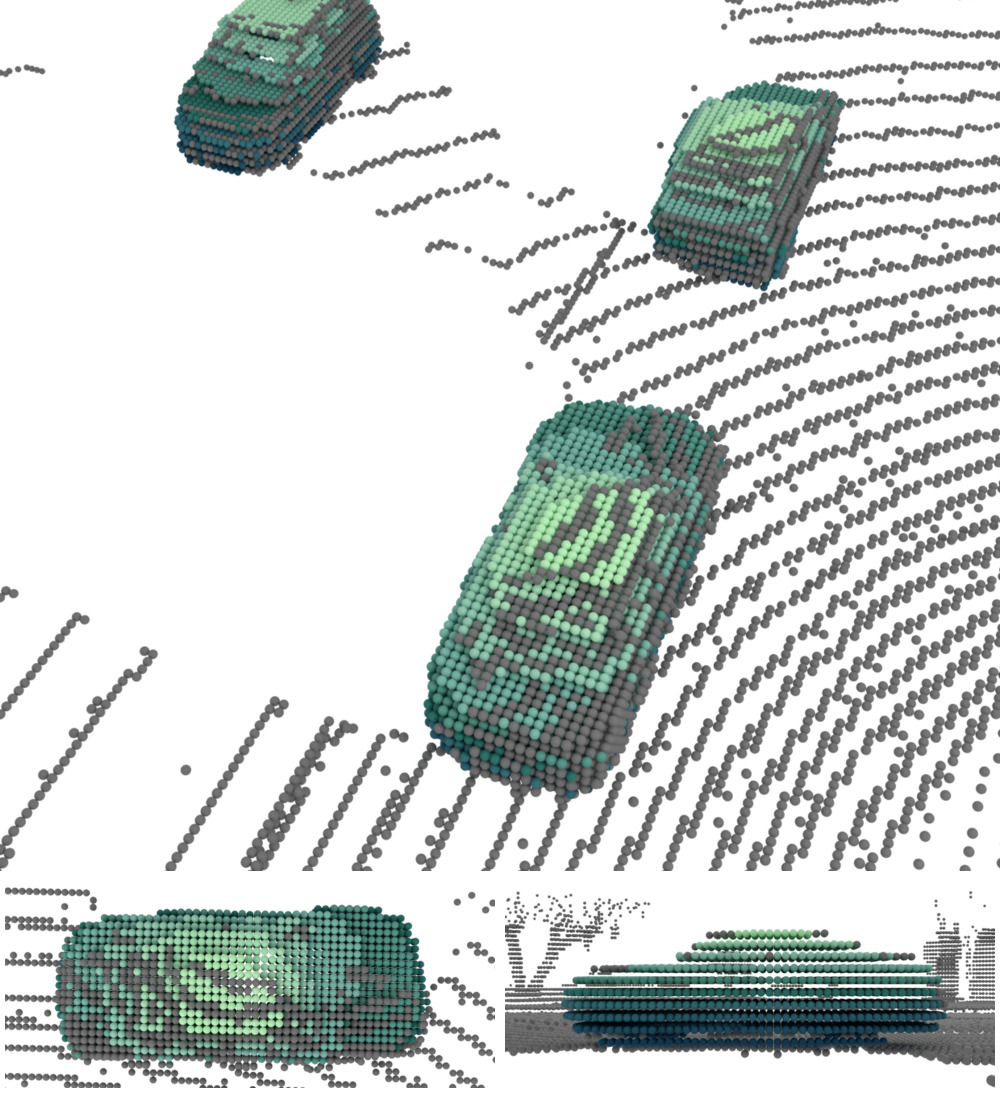}
	\caption{
		Reconstructed point cloud. 
	In contrast to existing \gls{ssrl} approaches which simply reconstruct the initial point cloud (gray), MAELi learns an expressive feature representation which captures the full geometric object structure of objects, without any ground truth labels.
    For visualization purposes, we color-coded the points by their $z$-coordinate and removed the reconstructed ground plane.
	}
	\label{fig:teaser}
\end{figure}

\glsreset{ssrl}
\Gls{ssrl} provides a technique to reduce the costly labeling effort.
The overall idea is to learn a universal feature representation in an unsupervised way, that is later utilized for a specific downstream task, such as object detection.
One of the most common approaches in 3D is to learn representations via \emph{point cloud reconstruction}~\cite{achlioptasLearningRepresentationsGenerative2018, wangOcCo2021, wenPointCloudCompletion2020, yuPointBERTPreTraining3D2022, pangMaskedAutoencodersPoint2022, liuMaskedDiscriminationSelfSupervised2022, tianGeoMAEMaskedGeometric2023}. 
There, the task is to restore removed parts of a point cloud and thereby learning an implicit understanding of the scene and the object's geometric structure. 
This is especially useful for full 3D point clouds generated from CAD models, such as ModelNet~\cite{wu3DShapeNetsDeep2015} or ShapeNet~\cite{changShapeNetInformationRich3D2015}.
Recently, these methods were adapted for large-scale point clouds within the automotive domain~\cite{minOccupancyMAE2023, hessMaskedAutoencoderSelfSupervised2023, tianGeoMAEMaskedGeometric2023}. 

Existing SSRL approaches, however, neglect inherent, but fundamental properties of LiDAR point clouds: i) We cannot sense beyond a hit surface (\ie we are limited to 2.5D perception), and ii) LiDAR sensors have a limited angular resolution.
In this work, we adapt to these properties and propose \emph{MAELi}, a transformer-less \gls{mae}, which does not simply follow the straightforward methodology of reconstructing the original LiDAR point cloud.
Instead, we present a novel reconstruction approach that allows us to go beyond the (visible) points. 
As a result, MAELi learns how objects look like from any viewing direction, which leads to strong pre-training weights with favorable generalization capabilities.
As illustrated in \Cref{fig:teaser}, it implicitly learns to reconstruct entire objects whilst training on single frames and in a genuinely unsupervised way without any ground truth whatsoever.

Intuitively, we explicitly distinguish \emph{occupied}, \emph{empty} and \emph{unknown} space. 
When tracing the path of a LiDAR beam from the sensor to an object (and back), the intervening space is considered \emph{emtpy}, the object itself is \emph{occupying} space at the point of impact and the space behind the object is \emph{unknown} due to occlusion. 
Moreover, no conclusions can be made about the regions that were not covered due to the limited resolution of the LiDAR.
Thus, we task our model with reconstructing removed parts of the point cloud, but we do not penalize it for the completion of structures in \emph{unknown} regions, \ie occluded or unsampled areas.

During training, the model encounters a wide range of objects, differing in pose and sampling density.
Despite the missing labels, our unique objective allows the model to reconstruct whole objects, \ie implicitly capturing the entire geometric structure.
In this process, MAELi learns a representation that more closely align with the underlying geometric structure, rather than simply mimicking the specific sampling patterns observable in a LiDAR point cloud.

For evaluation, we select the most predominant automotive perception tasks, \ie object detection and semantic segmentation. Our memory-efficient, sparse decoder structure enables efficient pre-training of state-of-the-art object detectors in an end-to-end fashion on a single GPU.
In extensive experiments on the Waymo Open Dataset~\cite{sunScalabilityPerceptionAutonomous2020}, KITTI~\cite{geigerAreWeReady2012, liaoKITTI360NovelDataset2023, behleySemanticKITTIDatasetSemantic2019} and ONCE~\cite{maoOneMillionScenes2021}, we demonstrate that MAELi is highly effective for pre-training various state-of-the-art 3D detectors and semantic segmentation networks.

In summary, our contributions are threefold:
\begin{itemize}
	\item We propose a LiDAR-aware \gls{ssrl} approach to pre-train 3D backbones applicable to various architectures and downstream tasks.
	\item We introduce a novel masking strategy and reconstruction loss for unsupervised representation learning, especially designed for LiDAR properties. 
    \item We show the effectivity of our pre-trained, visually verifiable representation improving several baselines for 3D object detection and semantic segmentation.
\end{itemize}

\section{Related Work}
\label{sec:rel_work}

\begin{figure*}
	\centering
	\includegraphics[width=\linewidth]{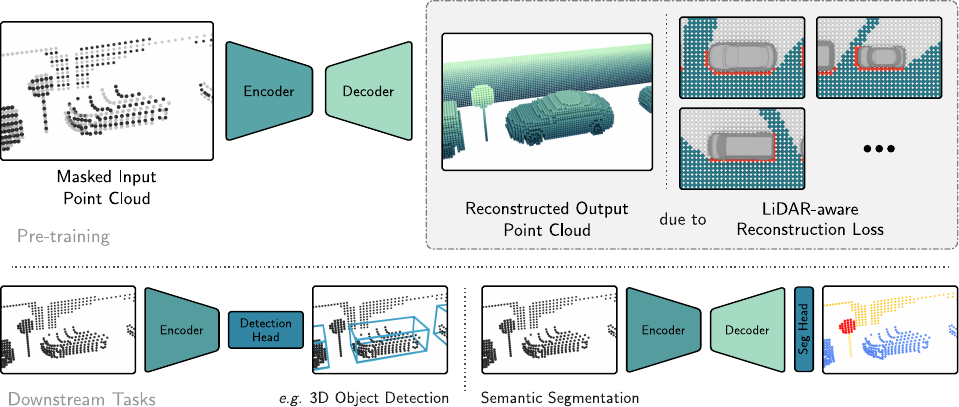}
	\caption{Schematic overview of our MAELi pre-training. The task of our sparse decoder is to reconstruct the missing parts of the masked input point cloud. 
			Thereby, the encoder is forced to learn a reasonable representation usable for a downstream task, \eg 3D object detection. 
			Since we do not penalize our network for reconstructing voxels at areas not visible to the LiDAR, after seeing numerous different samples, it learns to reconstruct occluded parts without any ground truth labels, leading to a more expressive feature representation. 
			The ground plane has been removed for visualization purposes.}
	\label{fig:overview}
	\vspace{-0.4cm}
\end{figure*}

\noindent\textbf{\gls{ssrl} for 2D Imagery and 3D Points Clouds:} Self-supervised representation learning strives to learn beneficial representations before introducing any supervision in the form of manually labeled ground truth data. 
These representations are utilized to improve the results on respective downstream tasks or to reduce the required amount of labeled training data. 

\textit{Contrastive learning} approaches task a model to maintain similar embeddings for the same data instance when transformed with different augmentations. 
Consequently, different data instances should lead to diverging embeddings. 
Initially applied to 2D imagery~\cite{hjelmLearningDeepRepresentations2019, chenSimpleFrameworkContrastive2020, heMomentumContrastUnsupervised2020, wangExploringCrossImagePixel2021}, these methods were also adapted for point clouds~\cite{zhangSelfSupervisedPretraining3D2021, xiePointContrastUnsupervisedPretraining2020, liangExploringGeometryAwareContrast2021, huangSpatioTemporalSelfSupervisedRepresentation2021, yinProposalContrastUnsupervisedPretraining2022, shengContrastivePredictiveAutoencoders2023, shengContrastivePredictiveAutoencoders2023, nunesSegContrast3DPoint2022, nunesTemporalConsistent3D2023, wuMaskedSceneContrast2023}.
Naturally, the granularity of the induced consistency loss defines the \emph{semantic level} on which the model agrees upon. 
In other words, contrastive learning on a global embedding describing an entire image or point cloud is more suitable for downstream tasks like classification. 
Tasks like object detection or semantic segmentation, however, necessitate a more fine-grained treatment. The causality dilemma is to sample semantically coherent regions with a proper level of detail without knowing what is semantically coherent. 
For example, Yin \etal~\cite{yinProposalContrastUnsupervisedPretraining2022} generate proposals via farthest point sampling and ball queries after removing the ground plane.  
TARL~\cite{nunesTemporalConsistent3D2023} and STSSL~\cite{wuSpatiotemporalSelfSupervisedLearning2023} extend the analysis to multiple time frames to cluster objects of interest, requiring the use of globally registered point clouds. 
This added temporal dimension introduces an additional layer of information, distinguishing them from single-frame based methods.

\noindent\textbf{\gls{ssrl} via Reconstruction:}
Recently, generative self-supervised representation learning approaches have been on the rise. 
One of its most successful concepts is the \emph{denoising autoencoders}. 
Based on the encoder's output embedding, the decoder is tasked to reconstruct the denoised input and if successful, the encoder is forced to learn a useful representation resilient to noise. 
Especially the reconstruction of a masked input gained huge traction across various domains of application, among others like \gls{nlp}~\cite{devlinBERTPretrainingDeep2019} and 2D imagery~\cite{heMaskedAutoencodersAre2022} also on point clouds. 
The predominant research focus started on learning representations on full 3D, synthetic or indoor datasets~\cite{girdharLearningPredictableGenerative2016, yangProgressiveSeedGeneration2021, wangOcCo2021, pangMaskedAutoencodersPoint2022, yuPointBERTPreTraining3D2022, liuMaskedDiscriminationSelfSupervised2022, yanImplicitAutoencoderPoint2022, zhangPointM2AEMultiscaleMasked2022, chenPiMAEPointCloud2023, houMask3DPreTraining2D2023}, \eg ModelNet~\cite{wu3DShapeNetsDeep2015} or ScanNet~\cite{daiScanNetRichlyAnnotated3D2017}. 

Recently, a few works consider LiDAR point clouds, such as~\cite{hessMaskedAutoencoderSelfSupervised2023, minOccupancyMAE2023, xieMaskedAutoencoderPreTraining2022, tianGeoMAEMaskedGeometric2023, yangGDMAEGenerativeDecoder2023, xuMVJARMaskedVoxel2023}. 
Xie~\etal~\cite{xieMaskedAutoencoderPreTraining2022} require full supervision, whereas we do not need any labels for pre-training.
MV-JAR~\cite{xuMVJARMaskedVoxel2023} tasks a backbone to reconstruct voxel's masked positional encoding and inner point distribution, enforcing the network to reconstruct the exact sampling pattern.
In Occupancy-MAE~\cite{minOccupancyMAE2023}, the authors use an MAE-approach to pre-train common 3D voxel backbones without the \gls{bev} encoder,
while GD-MAE~\cite{yangGDMAEGenerativeDecoder2023} introduces a newly developed elaborate multi-stage transformer.
Both approaches employ a dense decoder and do not differentiate between the empty and occluded spaces inherent in a LiDAR point cloud.
In contrast, our reconstruction strategy combined with our sparse decoder enables us to pre-train the entire encoder of the most common 3D object detectors without penalizing reconstruction in unsampled areas on a single GPU.

Only a few studies consider the inherent properties of a LiDAR point cloud, \ie its sampling resolution and 2.5D perception.
In this context, Wang \etal~\cite{wangOcCo2021} synthesize occlusions on small-scale full 3D point scans and leverage this to pre-train an encoder.
Hu \etal~\cite{huWhatYouSee2020} generate visibility maps via raycasting to mitigate inconsistent object augmentation~\cite{yanSECONDSparselyEmbedded2018}, serving as an additional input to a detection network.
Xu \etal~\cite{xuCurtainLearningOccluded2022} enhance ground truth object points by grouping similar instances and utilize them to train an auxiliary task, estimating the likelihood of an occluded area being occupied by an object.
GeoMAE~\cite{tianGeoMAEMaskedGeometric2023} proposes a masked autoencoder, which is trained to reconstruct underlying point statistics and surface properties, \ie an estimation of normal and curvature based on neighboring voxels.
ALSO~\cite{boulchALSOAutomotiveLidar2023}, most closely related to our approach, employs \gls{ssrl} via surface interpolation. 
It generates query points along LiDAR rays situated in front and behind the detected hits and instructs the network to predict the occupancy of these selected points, treating occluded areas as occupied. 
While this strategy is straightforward, it may encounter limitations in a multi-LiDAR setup on the ego vehicle, a configuration notably utilized in the Waymo Open Dataset~\cite{sunScalabilityPerceptionAutonomous2020}. 
In comparison, our approach is capable to handle such complex scenarios robustly and can implicitly learn geometric structure beyond sampled query points.
The self-supervision tasks of both, GeoMAE and ALSO, are deemed achieved once the network is able to interpolate the underlying surface. 
While the resulting feature representation already shows promising results, our LiDAR-aware reconstruction demonstrates further improvements as our model inherently captures the whole object geometry.

\section{MAE for Large-scale LiDAR Point Clouds}
\label{sec:method}

\glsreset{ssrl}
We aim to significantly reduce the expensive labeling effort for large-scale LiDAR point clouds via \gls{ssrl}.
We build upon the successful masking and reconstruction paradigm, but address the fundamental limitation that existing approaches primarily focus on reconstructing the original input point cloud.
While this strategy is effective for full 3D models, such as those generated by CAD renderings~\cite{wu3DShapeNetsDeep2015,changShapeNetInformationRich3D2015}, it limits the usefulness of the learned representation for LiDAR point clouds in two principal aspects.

First, the limited angular resolution of a LiDAR sensor induces gaps between the LiDAR beams.
Simply reconstructing the original point cloud would mean to penalize the model for (correctly) reconstructed points in these gaps.
Second, a single LiDAR sweep cannot capture objects fully.
Once a beam is reflected by a surface, the sensor is unable to capture any spatial information from objects situated behind that surface.
Thus, models based on standard reconstruction are penalized for completing occluded parts and, as a consequence, hindered from inherently understanding the underlying objects and learning implicit contextual information.
With MAELi, we address these challenges by introducing  our \emph{LiDAR-aware loss}. 
Rather than penalizing uniformly, this loss specifically targets known regions sampled by the LiDAR. 

While our approach is versatile and can be applied to a wide range of tasks, we chose to demonstrate its efficacy specifically on 3D detection and semantic segmentation due to their paramount importance and widespread application in the field.
For 3D detection, we pre-train the encoder by attaching a reconstruction decoder to its final layer.
Once pre-training is complete, we discard the decoder and utilize the encoder's weights as an initialization for the subsequent detection task.
For semantic segmentation, we use the weights of both the encoder and decoder as an initialization for downstream fine-tuning.
This is illustrated in \Cref{fig:overview}.

In the following, we briefly explain our sparse decoder (Sec.~\ref{subsec:sparse-decoder}), before describing our reconstruction objective (Sec.~\ref{subsec:objective}) and masking strategy (Sec.~\ref{subsec:masking}).\\

\noindent\textbf{Definitions \& Notations: }
Utilizing sparse operations, a voxel position is considered \emph{active} if any of its corresponding features deviates from zero and is consequently included in the computation process.
Only the active sites are actually stored in memory. 
Furthermore, we follow the common notation, \eg~\cite{choy4DSpatioTemporalConvNets2019a}, for sparse convolutions, where the \emph{voxel/tensor stride} $\mathbf{s}$ refers to the distance between two voxels w.r.t.~the highest voxel resolution along each axis.
For example, applying two (or three) 3D convolutions with a stride of $\mathbf{2}$ leads to a feature map with a tensor stride of $\mathbf{4}$ (or $\mathbf{8}$).
A downsampling step increases the stride, while an upsampling step decreases it. 

So, let $\mathcal{V}^{E,\mathbf{s}} = \{\mathbf{v}_{i}^{E,\mathbf{s}}\}_{i=1 \ldots M^{E,\mathbf{s}}}$ be the $M^{E,\mathbf{s}}$ \emph{active} voxel positions of a certain \emph{tensor stride}~$\mathbf{s}$ in the sparse encoder and $\mathcal{V}^{D,\mathbf{s}} = \{\mathbf{v}_{i}^{D,\mathbf{s}}\}_{i=1 \ldots M^{D,\mathbf{s}}}$ the $M^{D,\mathbf{s}}$ active voxel positions in the sparse decoder, where ${\mathbf{v}_i \in \mathbb{R}^{3}}$. 
To avoid cluttering the notation, unless stated otherwise, $\mathcal{V}^{\mathbf{s}}$ and $\mathbf{v}_{i}^{\mathbf{s}}$ imply the decoder and default to $\mathcal{V}^{D,\mathbf{s}}$ and $\mathbf{v}_i^{D,\mathbf{s}}$, respectively.

\subsection{Sparse Reconstruction Decoder}
\label{subsec:sparse-decoder}

Our goal is to obtain more expressive feature representations by pre-training the entire backbone for the respective downstream task.
Contrary to existing sparse, voxel-based encoder/decoder structures for LiDAR point clouds, such as Part-A$^2$~\cite{shiPointsParts3D2021}, we must address a key difference.
The typical approach involves voxelizing and processing the sparse data via dedicated \glspl{sc}.
Unlike its dense counterpart, a \gls{sc} is only applied if the kernel covers any \emph{active} sites. Even with small $3 \times 3$ kernels, these active sites dilute rapidly, escalating computational effort.
Thus, methods like~\cite{yanSECONDSparselyEmbedded2018, yinCenterBased3DObject2021, shiPVRCNNPointVoxelFeature2020} use \glspl{ssc}~\cite{graham3DSemanticSegmentation2018}, where the kernel center is placed only on active sites, considering only active sites covered by the kernel while maintaining favorable memory consumption.
The decoder then re-uses the active sites of the respective encoder layer during upsampling.
This, however, renders the common upsampling schema ineffective for reconstructing voxels not present in the encoder, making it unsuitable for a reconstruction task.

\begin{figure*}
	\centering

	\includegraphics[width=\linewidth]{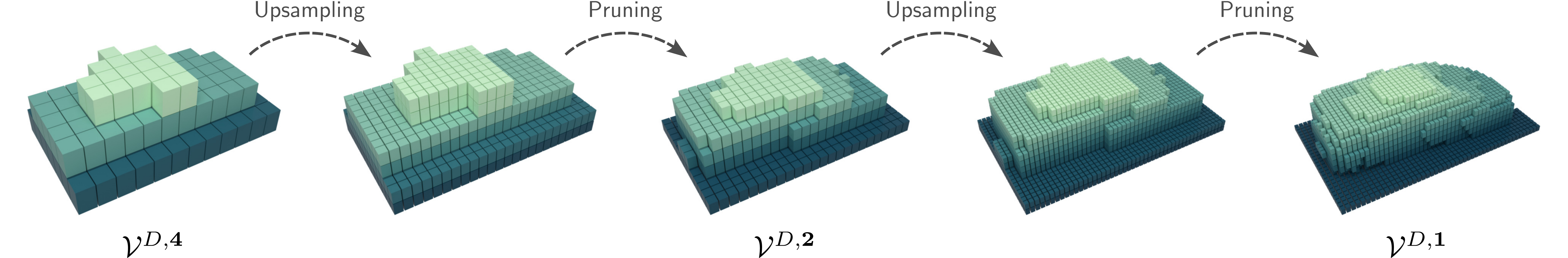}
	\caption{Reconstruction of an actual car within our decoder. An upsampling step halves the \emph{voxel stride} and voxel size and increases the number of voxels. Subsequently, superfluous voxels are removed during the pruning step. We cropped the car and color coded the $z$-coordinate for visualization purposes.}
	\label{fig:upsampling}
\end{figure*}

To perform reconstruction on a point cloud, we require a decoder that can also restore the removed voxels and expand beyond the active sites present in the encoder.
Dense upsampling to the original spatial resolution is a possibility, but it is rather impractical for pre-training larger architectures on limited hardware setups due to excessive memory and computational demands.
Thus, we allow each upsampling layer to grow (cubically), but add a subsequent pruning layer that learns to remove redundant voxels, as illustrated in \Cref{fig:upsampling}.
For this, we leverage the idea of small-scale point cloud reconstruction~\cite{choy4DSpatioTemporalConvNets2019a} and apply a $1 \times 1$ convolution to the sparse feature map that learns the pruning.
This way, the decoder is able to reconstruct and complete parts of the point cloud, while the pruning layer removes superfluous voxels.
To obtain stronger feature representations, we propose an extended reconstruction objective in the following.

\subsection{Reconstruction Objective}
\label{subsec:objective}

\begin{figure}[b]
	\centering
	\includegraphics[width=0.98\linewidth]{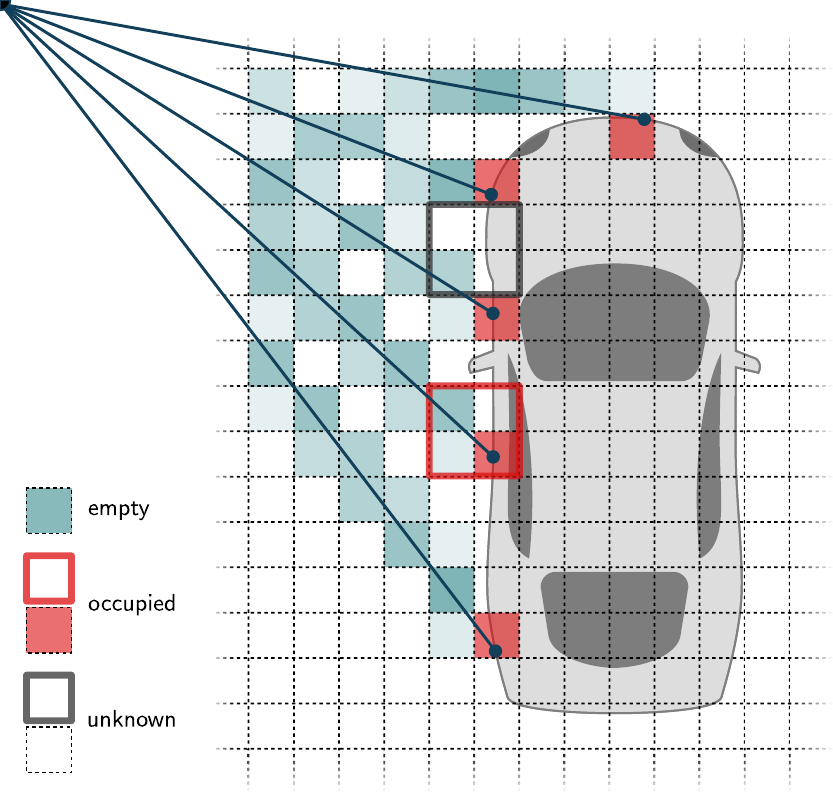}
	\caption{
	A loss should only be induced for voxels, where we \emph{actually know} whether the space is empty or not.
    LiDAR beams traverse \emph{empty} voxels (blue) until they hit a surface, \ie an \emph{occupied} voxel (red).
    All other voxels are considered \emph{unknown}. 
    The loss for an \emph{empty} voxel is weighted by the proximity of its center to the nearest beam (different shades of blue) to counter inaccuracies due to discrete voxelization.
    }
	\label{fig:loss}
\end{figure}

We formulate our reconstruction objective to incorporate a deeper understanding of the underlying environmental structure. 
Intuitively, it should be more straightforward for the model to grasp the complete appearance of a car rather than merely reconstructing specific points captured by LiDAR beams.
Moreover, the pre-trained representation should be more closely aligned with the objectives of downstream tasks, \eg fitting a 3D bounding box around an entire car. 
Consequently, during the reconstruction within the decoder, we distinguish three categories of voxels: \emph{occupied}, \emph{empty}, and \emph{unknown}. 
These categories are depicted in \Cref{fig:loss}.

Occupied voxels contain surface points from the original point cloud (before masking) and should thus be part of the reconstruction. 
Empty voxels are the ones traversed by the LiDAR beam without hitting any surface and thus, should remain empty.
Finally, we categorize a voxel as \emph{unknown} if neither of the first two cases applies, \ie these are either occluded or were not sensed by the beam due to the limited angular resolution.
This categorization enables the reconstruction to grow beyond the initially sensed point cloud.
Intuitively this makes sense, since it is counter-productive to punish the network for reconstructing points not sampled by the LiDAR, even if they are part of the underlying object.

We observe that the discrete voxelization can cause inaccuracies, especially at low resolution voxel grids.
To mitigate this, we reduce the loss for an empty voxel via the perpendicular distance $d_i^{\mathbf{s}}$ from the voxel center $\mathbf{v}_{i}^{\mathbf{s}}$ to the closest LiDAR beam.
More formally, we define the weight $w_i^{\mathbf{s}}$ for a voxel as
\begin{equation}	
	w_i^{\mathbf{s}} = \begin{cases}
		0 & \text{if $\mathbf{v}_{i}^{\mathbf{s}}$ is \emph{unknown}},\\
		1 & \text{if $\mathbf{v}_{i}^{\mathbf{s}}$ is \emph{occupied}},\\
		1 - \frac{2\,d_i^{\mathbf{s}}}{d_v^{\mathbf{s}}} & \text{if $\mathbf{v}_{i}^{\mathbf{s}}$ is \emph{empty}},
	\end{cases}
\end{equation}
where $d_v^{\mathbf{s}}$ denotes the length of a voxel diagonal at stride~$\mathbf{s}$. 
Note that a setup with multiple LiDARs (\eg Waymo~\cite{sunScalabilityPerceptionAutonomous2020}) can simply be handled by iterating the categorization process for each LiDAR.
There, the number of \emph{unknown} voxels would be reduced due to the additional LiDAR sensors.

We require the same categorization for voxels of lower resolution and larger strides, respectively.
For example, a voxel of stride $\mathbf{s}$ spatially covers eight voxels of stride $\mathbf{s}/2$ after upsampling (see \Cref{fig:upsampling}). 
Here, however, we need to consider that if a low-resolution voxel is pruned, it can no longer generate a self-supervision signal for its higher-resolution voxels during upsampling.
Thus, a low-resolution voxel is considered \emph{occupied} if it incorporates any high-resolution \emph{occupied} voxel (red square in \Cref{fig:loss}).
Similarly, it is \emph{unknown} (gray square) if it incorporates any \emph{unknown} voxel but no \emph{occupied} ones.
Otherwise, the low-resolution voxel is categorized as \emph{empty}.

In summary, the objective during self-supervised pre-training is to correctly classify whether a voxel is \emph{occupied} or \emph{empty}, while not penalizing \emph{unknown} ones.
For this, we use a weighted binary cross-entropy loss
\begin{equation}
	\mathcal{L} = - \frac{1}{\widetilde{M}}\sum_{\mathbf{s} \in \mathcal{S}} \sum_{i=1}^{M^{\mathbf{s}}} w_i^{\mathbf{s}}\,l_i^{\mathbf{s}}, \;\text{with}
\end{equation}
\begin{equation}
	l_i^{\mathbf{s}} = \left[ y_i^{\mathbf{s}} \log x_i^{\mathbf{s}} 
	+ (1 - y_i^{\mathbf{s}}) \log (1 - x_i^{\mathbf{s}}) \right],\\
\end{equation}
where $\mathcal{S}$ denotes the set of all strides; $\widetilde{M}$ is the amount of all \emph{occupied} and \emph{empty} voxels in the decoder; $y_i^{\mathbf{s}}$ is the actual occupancy of the voxelized input point cloud, \ie 1 if $\mathbf{v}_{i}^{\mathbf{s}}$ is \emph{occupied} and 0 otherwise; and $x_i^{\mathbf{s}}$ is our model's predicted occupancy probability.
In other words, we penalize each pruning step for removing \emph{occupied} and maintaining \emph{empty} voxels, but we do not induce any loss for \emph{unknown} voxels.

\subsection{Masking Strategy}
\label{subsec:masking}

In methods applied to small-scale full 3D point clouds, a common masking strategy is to apply \gls{fps} and the \acrlong{knn} algorithm to create overlapping patches~\cite{yuPointBERTPreTraining3D2022, pangMaskedAutoencodersPoint2022}, which are then randomly removed. 
Applying \gls{fps} to LiDAR point clouds would, however, predominantly pick isolated points, which are usually single outliers far off.
Masking these would not be beneficial to learn a strong feature representation, as such outliers usually do not correspond to real scene structures and have only few neighboring points.
Instead, we exploit the voxelization step already in place and randomly mask voxels.

Furthermore, consider the sampling resolution of a standard spinning LiDAR sensor: It sends out multiple beams, waits for their return and then derives the distance to a hit obstacle, \ie the \emph{range}, from the elapsed time. 
Every beam has an inclination and rotates around a common vertical axis, defining an azimuth. 
Since two specific beams enclose a constant angle, the spatial resolution and thus, the number of points decreases with the distance to the hit object.
Considering \gls{ssrl} via point cloud reconstruction, there is less self-supervision with increasing distance to the sensor.

Intuitively, if we remove points from nearby regions, such that they look similar to sparser regions further away, the model should be able to better generalize to these.
We incorporate this \emph{spherical masking} idea by reducing the angular resolution in both, azimuth and inclination.
To do this efficiently, we subsample the LiDAR's range image.
In particular, we randomly sample two integers $1 \leq m_r, m_c \leq 4$ and filter all rows $r$ and columns $c$ of the range image for which $r \mod m_r \neq 0$ or $c \mod m_c \neq 0$, respectively. 
For example, using only every second row and column halves the angular resolution.  
This way, an object is sampled as if it would have been further away, but we are able to induce a stronger self-supervision signal during our reconstruction task.

\section{Experiments}
\label{sec:experiments}

We demonstrate our MAELi pre-training by focusing on key tasks in automotive perception, namely object detection (\Cref{subsec:object_detection}) and semantic segmentation (\Cref{subsec:semantic_segmentation}). 
To validate our approach, we employ a range of architectures and pre-train them on widely-recognized datasets in these domains. 
Specifically, we use the \acrlong{wod}~\cite{sunScalabilityPerceptionAutonomous2020}, KITTI~\cite{geigerAreWeReady2012,behleySemanticKITTIDatasetSemantic2019,liaoKITTI360NovelDataset2023} and ONCE~\cite{maoOneMillionScenes2021} datasets. 

We compare against state-of-the-art approaches and, unless stated otherwise, report their official results, taken either from official benchmarks or the corresponding publications.
An additional ablation study to isolate the contributions of various components in our methodology is provided in the supplementary material.

\noindent\textbf{Implementation Details:}
We integrate MAELi into the OpenPCDet\cite{openpcdet} framework (v0.5.2) and employ the Minkowski Engine~\cite{choy4DSpatioTemporalConvNets2019a} to design our sparse decoder.
We use a voxel size of $[0.05, 0.05, 0.1]$, $[0.1, 0.1, 0.15]$ and $[0.1, 0.1, 0.2]$ for KITTI, Waymo and ONCE, respectively.
Given the respective architecture and dataset, we pre-train for 30 epochs without any labeled ground truth.
We use Adam~\cite{kingmaAdamMethodStochastic2015} and a one-cycle policy~\cite{smithDisciplinedApproachNeural2018} with maximum learning rate of 0.003.
We conducted all experiments on a single NVIDIA\textsuperscript{\textregistered} Quadro RTX\textsuperscript{\texttrademark} 8000.

\subsection{3D Object Detection}
\label{subsec:object_detection}

Our pre-training methodology MAELi is well-suited for 3D object detectors as our loss formulation lets the model learn how an object should look like.
These pre-trained weights generalize well across datasets and allow for data-efficient fine-tuning of a detector on a target dataset.

For 3D detection, we design a sparse decoder with four blocks, each inverting one downsampling operation of the encoder (exact architecture in the supplementary material). 
We pre-train the entire encoder including the dense \gls{bev} backbone~\cite{yanSECONDSparselyEmbedded2018}. 
Therefore, to utilize the sparse processing capabilities of the decoder, we take the active voxels from the last layer of the sparse 3D encoder and sample them from the \gls{bev} feature map.
To fine-tune the different 3D detectors with our pre-trained weights, we warm-up the detection head by freezing the encoder for one epoch and train end-to-end according to the default OpenPCDet configuration. 

OpenPCDet provides several augmentation techniques, \ie random flip/rotation/scaling and \emph{ground truth sampling}~\cite{yanSECONDSparselyEmbedded2018}.
During pre-training, we utilize random flip/rotation/scaling in addition to our masking strategy (see \Cref{subsec:masking}).
For fine-tuning, we adopt all of them as-is, except \emph{ground truth sampling}~\cite{yanSECONDSparselyEmbedded2018} during the data efficiency experiments. 
We cannot apply its vanilla form there, as it samples ground truth objects across the entire dataset and copy-pastes these randomly into frames which contain only few objects.
Without change, this would add ground truth objects from other than our subsampled frames and thus, invalidate these studies.
Thus, we filter the original ground truth database and ensure that it only contains objects from the actually used frames. 

\begin{table}
	\scriptsize
	\centering
		\scriptsize
		\centering
		\begin{tabular}{p{1.7cm}|cccccccccc}						
			\toprule
			\textbf{Method} & \textbf{Pre-train} & \textbf{mAP} & \textbf{Car} & \textbf{Pedestrian} & \textbf{Cyclist} \\
			\midrule
			SECOND~\cite{yanSECONDSparselyEmbedded2018} & - & 65.35 & 81.50 & 48.82 & 65.72 \\
			+ ALSO~\cite{boulchALSOAutomotiveLidar2023} & nuScenes & 68.07 & 81.78 & 54.24 & 68.19 \\
			+ ALSO~\cite{boulchALSOAutomotiveLidar2023} & KITTI 3D & 67.68 & \cellcolor{petrol!60}81.97 & 51.93 & 69.14 \\
			+ ALSO~\cite{boulchALSOAutomotiveLidar2023} & KITTI360 & 68.31 & \cellcolor{petrol!25}81.79 & 52.45 & 70.68 \\
			\textbf{+ MAELi} & Waymo & \cellcolor{petrol!25}69.05 & 81.70 & 54.34 & \cellcolor{petrol!25}71.11 \\
			\textbf{+ MAELi} & KITTI 3D & 68.18 & 81.37 & \cellcolor{petrol!25}54.37 & 68.80 \\
			\textbf{+ MAELi} & KITTI360 & \cellcolor{petrol!60}69.51 & 81.51 & \cellcolor{petrol!60}54.74 & \cellcolor{petrol!60}72.28 \\
			\hhline{*{7}{-}}
			PV-RCNN~\cite{shiPVRCNNPointVoxelFeature2020} & - & 70.57 & 84.50 & 57.06 & 70.14 \\
			+ STRL~\cite{huangSpatioTemporalSelfSupervisedRepresentation2021} & KITTI 3D & 71.46 & 84.70 & 57.80 & 71.88 \\
			+ GCC-3D~\cite{liangExploringGeometryAwareContrast2021} & Waymo & 71.26 & - & - & - \\
			+ PropCont~\cite{yinProposalContrastUnsupervisedPretraining2022} & Waymo & 72.92 & 84.72 & \cellcolor{petrol!25}60.36 & 73.69 \\
			+ ALSO~\cite{boulchALSOAutomotiveLidar2023} & nuScenes & 72.53 & \cellcolor{petrol!60}84.86 & 57.76 & 74.98 \\
			+ ALSO~\cite{boulchALSOAutomotiveLidar2023} & KITTI 3D & 72.76 & 84.72 & 58.49 & \cellcolor{petrol!25}75.06 \\
			+ ALSO~\cite{boulchALSOAutomotiveLidar2023} & KITTI360 & \cellcolor{petrol!25}72.96 & 84.68 & 60.16 & 74.04 \\
			\textbf{+ MAELi} & Waymo & 72.15 & \cellcolor{petrol!25}84.80 & 57.46 & 74.18 \\
			\textbf{+ MAELi} & KITTI 3D & 72.58 & 82.94 & 59.51 & \cellcolor{petrol!60}75.29 \\
			\textbf{+ MAELi} & KITTI360 & \cellcolor{petrol!60}73.43 & 84.71 & \cellcolor{petrol!60}62.29 & 73.27 \\
			\bottomrule		
		\end{tabular}	
		\caption{
			Performance comparison of pre-trained weights fine-tuned on the full KITTI 3D training set for SECOND and PV-RCNN.
			Results are reported on the KITTI 3D \emph{val} set using the standard $R_{40}$ metric.
		}
		\label{tab:kitti_full_r40}
\end{table}

\paragraph{Detection Results:}
The \textbf{KITTI 3D} dataset and benchmark~\cite{geigerAreWeReady2012} is one of the first publicly available datasets for 3D object detection and consists of $\sim7.5$k LiDAR frames, which were labeled in the front camera view. 
KITTI360~\cite{liaoKITTI360NovelDataset2023} is an extension of the original KITTI dataset, containing $\sim80$k LiDAR frames, offering additional modalities and more exhaustive annotations.
For a comparison to other pre-training approaches, we show detection results on the KITTI 3D set while using pre-trained weights from KITTI 3D, KITTI360 and Waymo. 
For our experiments, we report evaluation metrics based on the moderate difficulty level, utilizing the official $R_{40}$ metric with 40 recall points.

\Cref{tab:kitti_full_r40} shows the detection performance when using different pre-trained weights for the widely-used SECOND~\cite{yanSECONDSparselyEmbedded2018} and PV-RCNN~\cite{shiPVRCNNPointVoxelFeature2020} detectors.
Throughout all datasets used for pre-training, MAELi enables strong improvements over detection models trained from scratch and yields top performing detection results.
In particular, we outperform the current state-of-the-art, \ie ALSO, in overall mAP, while the performance difference in the car category remains within a very narrow margin, indicating a saturation in performance for this class.

The large-scale \textbf{\acrlong{wod}}~\cite{sunScalabilityPerceptionAutonomous2020} contains 798 training sequences and 202 validation sequences.
Labels are provided for vehicles, pedestrians and cyclists.
We report our results using the official Waymo evaluation protocol, reporting APH at the more challenging LEVEL 2 difficulty. 
Detailed results including the AP scores are provided in the supplemental material.

We pre-train our weights on the Waymo \emph{train} set and use them to initialize the backbones of SECOND~\cite{yanSECONDSparselyEmbedded2018}, CenterPoint~\cite{yinCenterBased3DObject2021} and PV-RCNN~\cite{shiPVRCNNPointVoxelFeature2020}.
For a fair comparison to~\cite{yinProposalContrastUnsupervisedPretraining2022, minOccupancyMAE2023}, we follow the common protocol of using 20\% of the entire Waymo \emph{train} set, including vanilla ground truth sampling, to fine-tune the detectors.

\Cref{tab:waymo_comparison} shows the Waymo results in comparison to the pre-training approaches Occupancy-MAE~\cite{minOccupancyMAE2023}, GCC-3D~\cite{liangExploringGeometryAwareContrast2021} and ProposalContrast~\cite{yinProposalContrastUnsupervisedPretraining2022}.
Pre-training with MAELi demonstrates strong improvements across all detectors.
While MAELi pre-trained weights are clearly better suited for SECOND than Occupancy-MAE's, the results for CenterPoint and PV-RCNN are on par with Occupancy-MAE, clearly outperforming GCC-3D or ProposalContrast.

\begin{table}
	\scriptsize
	\centering
	\begin{tabular}{p{1.7cm}|HcHcHcHcHc}
		\toprule
		\multirow{2}{*}{\textbf{Method}} &  \multicolumn{8}{c}{\textbf{3D APH (LEVEL 2)}}        \\
		&  \multicolumn{2}{c}{\textbf{Gain}}  & \multicolumn{2}{c}{\textbf{Overall}} & \multicolumn{2}{c}{\textbf{Vehicle}} & \multicolumn{2}{c}{\textbf{Pedestrian}} & \multicolumn{2}{c}{\textbf{Cyclist}} \\
		\midrule
		SECOND~\cite{yanSECONDSparselyEmbedded2018} & - & - &  58.26 & 54.35 &  62.58 & 62.02 &  57.22 & 47.49 &  54.97 & 53.53 \\
		+ Occ-MAE~\cite{minOccupancyMAE2023} & \cellcolor{petrol!25}+0.85 & \cellcolor{petrol!25}+0.75            &  \cellcolor{petrol!25}59.11 & \cellcolor{petrol!25}55.10 &  \cellcolor{petrol!25}62.67 & \cellcolor{petrol!25}62.34 &  \cellcolor{petrol!25}59.03 & \cellcolor{petrol!25}48.79 &  \cellcolor{petrol!25}55.62 & \cellcolor{petrol!25}54.17 \\
		\textbf{+ MAELi}  & \cellcolor{petrol!60}+2.32 & \cellcolor{petrol!60}+2.35 &  \cellcolor{petrol!60}60.57 & \cellcolor{petrol!60}56.69 &  \cellcolor{petrol!60}63.75 & \cellcolor{petrol!60}63.20 &  \cellcolor{petrol!60}60.71 & \cellcolor{petrol!60}50.93 &  \cellcolor{petrol!60}57.26 & \cellcolor{petrol!60}55.95 \\
		\hhline{*{11}{-}}
		CenterPoint~\cite{yinCenterBased3DObject2021}  & - & -             &  64.51 & 61.92 &  63.16 & 62.65 &  64.27 & 58.23 &  66.11 & 64.87 \\
		+ Occ-MAE~\cite{minOccupancyMAE2023} & \cellcolor{petrol!60}+1.35 & \cellcolor{petrol!60}+1.31 &  \cellcolor{petrol!60}65.86 & \cellcolor{petrol!60}63.23 &  \cellcolor{petrol!25}64.05 & \cellcolor{petrol!25}63.53 &  \cellcolor{petrol!25}65.78 & \cellcolor{petrol!25}59.62 &  \cellcolor{petrol!60}67.76 & \cellcolor{petrol!60}66.53 \\
		\textbf{+ MAELi} & \cellcolor{petrol!25}+1.09 & \cellcolor{petrol!25}+1.08     & \cellcolor{petrol!25}65.60 & \cellcolor{petrol!25}63.00 & \cellcolor{petrol!60}64.22 & \cellcolor{petrol!60}63.70 & \cellcolor{petrol!60}65.93 & \cellcolor{petrol!60}59.79 & \cellcolor{petrol!25}66.66 & \cellcolor{petrol!25}65.52 \\
		\hhline{*{11}{-}}
		PV-RCNN~\cite{shiPVRCNNPointVoxelFeature2020}  & - & - & 59.84 & 56.23 & 64.99 & 64.38 & 53.80 & 45.14 & 60.72 & 59.18 \\
		+ GCC-3D~\cite{liangExploringGeometryAwareContrast2021} & +1.46 & +1.95 & 61.30 & 58.18 & 65.65 & 65.10 & 55.54 & 48.02 & 62.72 & 61.43 \\
		+ PropCont~\cite{yinProposalContrastUnsupervisedPretraining2022} & +2.78 & +3.05 & 62.62 & 59.28 & 66.04 & 65.47 & 57.58 & 49.51 & \cellcolor{petrol!25}64.23 & \cellcolor{petrol!25}62.86 \\
		+ Occ-MAE~\cite{minOccupancyMAE2023} & \cellcolor{petrol!60}+5.99 & \cellcolor{petrol!25}+5.74 & \cellcolor{petrol!60}65.82 & \cellcolor{petrol!25}61.98 & \cellcolor{petrol!60}67.94 & \cellcolor{petrol!60}67.34 & \cellcolor{petrol!25}64.91 & \cellcolor{petrol!25}55.57 & \cellcolor{petrol!60}64.62 & \cellcolor{petrol!60}63.02 \\
		\textbf{+ MAELi} & \cellcolor{petrol!25}+5.88 & \cellcolor{petrol!60}+5.92 & \cellcolor{petrol!25}65.72 & \cellcolor{petrol!60}62.15 & \cellcolor{petrol!25}67.90 & \cellcolor{petrol!60}67.34 & \cellcolor{petrol!60}65.14 & \cellcolor{petrol!60}56.32 & 64.13 & 62.79 \\
		\bottomrule
	\end{tabular}\\	
	\caption{Performance comparison on the Waymo \emph{val} set trained on 20\% of the Waymo \emph{train} set. We compare different detectors trained from scratch with their pendants utilizing pre-trained weights from GCC-3D, ProposalContrast, Occupancy-MAE and the proposed MAELi.
	}
	\label{tab:waymo_comparison}
\end{table}

\textbf{ONCE}~\cite{maoOneMillionScenes2021} is a comprehensive dataset designed for tasks such as 3D object detection, tracking and motion forecasting.
This dataset includes 1 million frames, the majority of which are unlabeled.
A primary objective of ONCE is to serve as a foundation for research that leverages on large-scale unlabeled data.
In our work, we utilize the official $U_{\text{small}}$ subset for pre-training purposes, subsequently fine-tuning our models on the training set and conducting evaluations on the validation set. 

\Cref{tab:ft_once} shows the results comparing MAELi with other pre-training methods on ONCE.
Besides our favorable results, this evaluation also demonstrates the strong cross-domain generalization capabilities of our pre-training approach: for SECOND~\cite{yanSECONDSparselyEmbedded2018}, we outperform the state-of-the-art approaches even when using pre-trained weights from Waymo, \ie $55.84$ mAP (MAELi pre-trained on Waymo) versus $52.68$ (ALSO pre-trained on $U_{\text{small}}$).
Naturally, pre-training on ONCE itself further improves the results.

\begin{table}
	\scriptsize
	\centering
	\begin{tabular}{p{1.75cm}|ccccc}
		\toprule
		\multirow{2}{*}{\textbf{Method}} & \multirow{2}{*}{\textbf{Pre-train}} & \multicolumn{4}{c}{\textbf{Orientation-Aware AP}}        \\
		& & \textbf{mAP} & \textbf{Car} & \textbf{Pedestrian} & \textbf{Cyclist} \\
		\midrule
		SECOND~\cite{yanSECONDSparselyEmbedded2018} & - & 51.89 & 71.19 & 26.44 & 58.04 \\
		+ SwAV~\cite{caronUnsupervisedLearningVisual2020} & $U_{small}$ & 51.96 & 72.71 & 25.13 & 58.05 \\
		+ DeepCluster~\cite{caronDeepClusteringUnsupervised2018} & $U_{small}$ & 52.06 & 73.19 & 24.00 & 58.99 \\
		+ ALSO~\cite{boulchALSOAutomotiveLidar2023} & $U_{small}$ & 52.68 & 71.73 & 28.16 & 58.13 \\
		\textbf{+ MAELi} & $U_{small}$ & \cellcolor{petrol!60}57.39 & \cellcolor{petrol!25}75.73 & \cellcolor{petrol!60}34.83 & \cellcolor{petrol!60}61.62 \\
		\textbf{+ MAELi} & Waymo & \cellcolor{petrol!25}55.84 & \cellcolor{petrol!60}75.86 & \cellcolor{petrol!25}31.03 & \cellcolor{petrol!25}60.65 \\
		\hhline{*{6}{-}}
		CenterPoint~\cite{yinCenterBased3DObject2021} & - & 64.24 & 75.26 & 51.65 & 65.79 \\
		+ PropCont~\cite{yinProposalContrastUnsupervisedPretraining2022} & Waymo & \cellcolor{petrol!25}66.24 & \cellcolor{petrol!25}78.00 & \cellcolor{petrol!60}52.56 & \cellcolor{petrol!25}68.17 \\
		\textbf{+ MAELi} & Waymo & \cellcolor{petrol!60}66.72 & \cellcolor{petrol!60}80.09 & \cellcolor{petrol!25}51.87 & \cellcolor{petrol!60}68.21 \\
		\bottomrule
	\end{tabular}\\	
	\caption{Performance comparison on the ONCE validation set. Our initialization, even when pre-trained on a different dataset, helps outperforming state-of-the-art methods.}
	\label{tab:ft_once}
\end{table}

\paragraph{Data Efficiency:}

Pre-trained weights play a crucial role during detector initialization.
The goal of self-supervised representation learning approaches is to reduce the costly labeling effort. If properly pre-trained, a detector should achieve strong performance with only few annotated samples.
We conduct the following experiments to demonstrate the benefits of MAELi for low-data regimes:
 
On \textbf{Waymo}, we follow the evaluation of ProficientTeachers~\cite{yinSemisupervised3DObject2022}, a state-of-the-art (but semi-supervised) approach which is specifically tailored for data-efficient 3D object detection.
The protocol is to pre-train SECOND using the first 399 Waymo \emph{train} sequences and then using different fractions of labeled data from the latter 399 sequences to fine-tune the detection heads.
For the evaluation on \textbf{KITTI 3D}, we adhere to the methodology presented in ProposalContrast~\cite{yinProposalContrastUnsupervisedPretraining2022}.
Thus, we pre-train PV-RCNN on the Waymo dataset and subsequently fine-tune it on various fractions of labeled KITTI 3D frames.

Tables~\ref{tab:data_efficiency_second} and~\ref{tab:data_efficiency_kitti} show the data efficiency evaluations for Waymo and KITTI, respectively.
Comparisons for other detectors are provided in the supplementary material.
Overall, MAELi performs favorably across the different low-data setups.
As with any SSRL approach, the initial benefits diminish as the downstream model is trained on larger quantities of annotated samples.
However, MAELi performs favorably with fewer annotations, where SSRL is most important since such techniques are designed to provide strong pre-trained weights for the crucial initialization phase.

\begin{table}
	\scriptsize
	\centering
	\setlength{\tabcolsep}{3pt}
	 \begin{tabular}{cp{1.65cm}|HcHcHcHcHc}						
        \toprule			
        \multirow{2}{*}{\textbf{Fraction}} & \multirow{2}{*}{\textbf{Method}} &  \multicolumn{10}{c}{\textbf{3D APH (LEVEL 2)}}        \\
        &  & \multicolumn{2}{c}{\textbf{Gain}} & \multicolumn{2}{c}{\textbf{Overall}} & \multicolumn{2}{c}{\textbf{Vehicle}} &  \multicolumn{2}{c}{\textbf{Ped.}} &  \multicolumn{2}{c}{\textbf{Cyc.}}              \\
        \midrule
        \multirow{2}{*}{\shortstack{1\%\\(791 frames)}} 
        & SECOND~\cite{yanSECONDSparselyEmbedded2018} &   - & - &           \cellcolor{petrol!25}31.09 & \cellcolor{petrol!25}22.25 &  \cellcolor{petrol!25}41.64 & \cellcolor{petrol!25}40.02 &  \cellcolor{petrol!25}33.39 & \cellcolor{petrol!25}17.45 &  \cellcolor{petrol!25}18.24 &  \cellcolor{petrol!25}\phantom{0}9.29 \\
        & \textbf{+ MAELi} & \cellcolor{petrol!60}+14.92 & \cellcolor{petrol!60}+10.79 &  \cellcolor{petrol!60}46.01 & \cellcolor{petrol!60}33.05 &  \cellcolor{petrol!60}51.05 & \cellcolor{petrol!60}50.11 &  \cellcolor{petrol!60}48.13 & \cellcolor{petrol!60}24.65 &  \cellcolor{petrol!60}38.86 & \cellcolor{petrol!60}24.38 \\
        \hhline{*{12}{-}}
        \multirow{3}{*}{\shortstack{5\%\\(3952 frames)}}
        & SECOND~\cite{yanSECONDSparselyEmbedded2018} &  - & - &  46.67 & 34.05 & 53.19 & 52.37 &  44.77 & 22.80 &  42.04 & 26.97 \\
        & + ProfTeach~\cite{yinSemisupervised3DObject2022} & \cellcolor{petrol!25}+4.43 & \cellcolor{petrol!25}+11.70 & \cellcolor{petrol!25}51.10 & \cellcolor{petrol!25}45.75 & 53.04 & \cellcolor{petrol!25}52.54 & \cellcolor{petrol!25}50.33 & \cellcolor{petrol!25}38.67 & \cellcolor{petrol!60}49.92 & \cellcolor{petrol!60}46.03 \\
        & \textbf{+ MAELi} &           \cellcolor{petrol!60}+6.29 & \cellcolor{petrol!60}+13.94 &  \cellcolor{petrol!60}52.96 & \cellcolor{petrol!60}47.99 &     \cellcolor{petrol!60}57.50 & \cellcolor{petrol!60}56.75 &  \cellcolor{petrol!60}53.45 & \cellcolor{petrol!60}41.27 &  \cellcolor{petrol!25}47.93 & \cellcolor{petrol!25}45.94 \\
        \hhline{*{12}{-}}
        \multirow{3}{*}{\shortstack{10\%\\(7904 frames)}}
        & SECOND~\cite{yanSECONDSparselyEmbedded2018} &            - & - &  52.74 & 38.23 &            \cellcolor{petrol!25}58.11 & \cellcolor{petrol!25}57.46 &  52.26 & 28.15 &  47.85 & 29.07 \\
        & + ProfTeach~\cite{yinSemisupervised3DObject2022} & \cellcolor{petrol!25}+2.27 & \cellcolor{petrol!25}+12.20  & \cellcolor{petrol!25}55.01 & \cellcolor{petrol!25}50.43  & 57.59 & 56.92 & \cellcolor{petrol!25}54.28 & \cellcolor{petrol!25}43.19 & \cellcolor{petrol!60}53.15 & \cellcolor{petrol!60}51.18 \\
        & \textbf{+ MAELi} &            \cellcolor{petrol!60}+3.23 &\cellcolor{petrol!60} +13.61 & \cellcolor{petrol!60}55.97 & \cellcolor{petrol!60}51.84 &     \cellcolor{petrol!60}60.13 & \cellcolor{petrol!60}59.47 &  \cellcolor{petrol!60}55.89 & \cellcolor{petrol!60}45.52 &  \cellcolor{petrol!25}51.90 & \cellcolor{petrol!25}50.52 \\
        \hhline{*{12}{-}}
        \multirow{3}{*}{\shortstack{20\%\\(15808 frames)}}
        & SECOND~\cite{yanSECONDSparselyEmbedded2018} &            - & - &   55.82 & 51.26 &           \cellcolor{petrol!25}60.16 & \cellcolor{petrol!25}59.54 &  54.28 & 43.30 &  53.03 & 50.93 \\
        & + ProfTeach~\cite{yinSemisupervised3DObject2022} & \cellcolor{petrol!60}+2.77 &\cellcolor{petrol!60} +2.90 & \cellcolor{petrol!60}58.59 & \cellcolor{petrol!60}54.16 & 59.97 & 59.36 & \cellcolor{petrol!25}57.88 & \cellcolor{petrol!25}46.97 & \cellcolor{petrol!60}57.93 & \cellcolor{petrol!60}56.15 \\
        & \textbf{+ MAELi} &           \cellcolor{petrol!25}+2.27 & \cellcolor{petrol!25}+2.75 &  \cellcolor{petrol!25}58.09 & \cellcolor{petrol!25}54.01 & \cellcolor{petrol!60}61.80 & \cellcolor{petrol!60}61.21 &  \cellcolor{petrol!60}57.91 & \cellcolor{petrol!60}47.63 &  \cellcolor{petrol!25}54.57 & \cellcolor{petrol!25}53.18 \\
        \bottomrule			
    \end{tabular}\\	
    \caption{        
        Data efficiency comparison for SECOND on the Waymo \emph{val} set.
        The first 399 Waymo \emph{train} sequences are used for pre-training and different fractions of the latter 399 sequences are used for fine-tuning.
    }
    \label{tab:data_efficiency_second}
\end{table}

\begin{table}
    \scriptsize
    \centering
    \begin{tabular}{cp{1.6cm}|ccccccccc}						
        \toprule
        \textbf{Fraction} & \textbf{Method} & \textbf{mAP} & \textbf{Car} & \textbf{Ped.} & \textbf{Cyc.} \\
        \midrule
        \multirow{3}{*}{\shortstack{20\%\\(743 frames)}} & PV-RCNN~\cite{shiPVRCNNPointVoxelFeature2020} & 66.71 &  \cellcolor{petrol!25}82.52 &      53.33 &   64.28 \\
        & + PropCont~\cite{yinProposalContrastUnsupervisedPretraining2022} & \cellcolor{petrol!25}68.13 & \cellcolor{petrol!60}82.65 & \cellcolor{petrol!25}55.05 & \cellcolor{petrol!25}66.68 \\
        & \textbf{+ MAELi} &   \cellcolor{petrol!60}69.41 & 82.21 & \cellcolor{petrol!60}56.71 & \cellcolor{petrol!60}69.30 \\
        \hhline{*{7}{-}}
        \multirow{3}{*}{\shortstack{50\%\\(1856 frames)}} & PV-RCNN~\cite{shiPVRCNNPointVoxelFeature2020} & 69.63 &  82.68 & \cellcolor{petrol!25}57.10 & 69.12 \\
        & + PropCont~\cite{yinProposalContrastUnsupervisedPretraining2022} &  \cellcolor{petrol!60}71.76 &  \cellcolor{petrol!25}82.92 & \cellcolor{petrol!60}59.92 & \cellcolor{petrol!60}72.45 \\
        & \textbf{+ MAELi} &  \cellcolor{petrol!25}70.13 & \cellcolor{petrol!60}83.89 & 56.48 & \cellcolor{petrol!25}70.02 \\
        \bottomrule		
    \end{tabular}
    \caption{
        Data efficiency comparison for PV-RCNN. Following ProposalContrast, we pre-train on Waymo and fine-tune on different fractions of KITTI 3D.
    }
    \label{tab:data_efficiency_kitti}
\end{table}

\subsection{Semantic Segmentation}
\label{subsec:semantic_segmentation}
Since our pre-training methodology does not depend on the downstream task, we further demonstrate its efficacy in the context of semantic segmentation.
To this end, we employ regular sparse transpose convolutions in the decoder, which are capable of expanding beyond the active voxels in the encoder, as opposed to using submanifold convolutions (see \Cref{subsec:sparse-decoder}).
Importantly, this maintains the weight dimensionality, enabling the direct utilization of pre-trained weights from both the encoder and decoder during the fine-tuning phase. 

We initialize with the pre-trained weights and employ the same framework and configurations as the state-of-the-art ALSO~\cite{boulchALSOAutomotiveLidar2023} for subsequent fine-tuning and evaluation procedures. 
Following ALSO, we use the MinkUNet~\cite{choy4DSpatioTemporalConvNets2019a} variant from~\cite{nunesSegContrast3DPoint2022} and report the average performance over 5 runs on the \textbf{SemanticKITTI} dataset~\cite{behleySemanticKITTIDatasetSemantic2019}.
This dataset provides semantic labels for each point cloud in the odometry task of the KITTI dataset~\cite{geigerAreWeReady2012}.
SemanticKITTI comprises 22 sequences, 19 classes and $\sim$23k frames for training and validation purposes. 
Our performance is assessed based on the official evaluation protocol, reported as the mean Intersection-over-Union~(mIoU) across all classes.

We follow the evaluation setup of ALSO: pre-training is conducted on the complete dataset and fine-tuning is done on varying fractions of the training set using the training and validation splits from~\cite{nunesSegContrast3DPoint2022}.
Our results, stated in \Cref{tab:ft_semantic_kitti}, reveal performance enhancements across almost all considered fractions of the training set (except the highly unreliable setting with merely 17 frames), thereby affirming the general applicability of our MAELi approach in semantic segmentation tasks.

\begin{table}
	\vspace{0.2cm}
	\scriptsize
	\centering
	\begin{tabular}{p{2.1cm}|ccccc}
		\toprule
		\multirow{3}{*}{\textbf{Method}} & \multicolumn{5}{c}{\textbf{Fraction of labeled samples}}        \\
		& \textbf{0.1\%} & \textbf{1\%} & \textbf{10\%} & \textbf{50\%} & \textbf{100\%} \\
		& \textbf{(17 fr.)} & \textbf{(188)} & \textbf{(1912)} & \textbf{(9560)} & \textbf{(19130)} \\
		\midrule
		MinkUNet~\cite{nunesSegContrast3DPoint2022} & 30.0 & 46.2 & 57.6 & 61.8 & 62.7 \\
		+ PointContrast~\cite{xiePointContrastUnsupervisedPretraining2020} & 32.4 & 47.9 & 59.7 & 62.7 & 63.4 \\
		+ DepthContrast~\cite{zhangSelfSupervisedPretraining3D2021} & 32.5 & 49.0 & 60.3 & 62.9 & \cellcolor{petrol!25}63.9 \\
		+ SegContrast~\cite{nunesSegContrast3DPoint2022} & 32.3 & 48.9 & 58.7 & 62.1 & 62.3 \\
		+ ALSO~\cite{boulchALSOAutomotiveLidar2023} & \cellcolor{petrol!60}35.0 & \cellcolor{petrol!25}50.0 & \cellcolor{petrol!25}60.5 & \cellcolor{petrol!25}63.4 & 63.6 \\
		\textbf{+ MAELi} & \cellcolor{petrol!25}34.6 & \cellcolor{petrol!60}50.7 & \cellcolor{petrol!60}61.3 & \cellcolor{petrol!60}63.6 & \cellcolor{petrol!60}64.2 \\
		\bottomrule
	\end{tabular}\\	
	\caption{
	Performance comparison on the SemanticKITTI dataset using a MinkUNet backbone.
	Results show the mean Intersection-over-Union (mIoU) averaged over 5 runs.
	}
	\label{tab:ft_semantic_kitti}
\end{table}

\section{Conclusion}
\label{sec:conclusion}

We proposed MAELi, a self-supervised pre-training approach, carefully designed to adapt to the inherent but subtle properties of large-scale LiDAR point clouds.
We were the first to put aspects like occlusion and intrinsic spherical sampling of LiDAR data into the context of \gls{ssrl}.
Our learned representation not only leads to significant improvements in various tasks, but can also be visually verified.
Moreover, it can be easily applied to other datasets, with minimal data requirements for fine-tuning.
Additionally, our method offers the potential for further investigation on a multi-frame basis.
With MAELi, we offer a new method to sustainably reduce the amount of tedious and costly annotation tasks for LiDAR point clouds.

{
\small
\noindent\textbf{Acknowledgements}
This work was partially supported by the Austrian Federal Ministry for Digital and Economic Affairs, the National Foundation for Research, Technology and Development and the Christian Doppler Research Association.
}

{\small
\bibliographystyle{ieee_fullname}
\bibliography{main}
}

\clearpage
\appendix
\part*{Supplementary Material} %
\hypersetup{linkcolor=black}
\setcounter{figure}{0}
\setcounter{table}{0}
\renewcommand{\thefigure}{S\arabic{figure}}
\renewcommand{\thetable}{S\arabic{table}}
This supplementary material presents further details, results and insights into MAELi.
We state further results in \Cref{sec:additional_results}, describe the detailed architecture of our decoder in \Cref{sec:architecture} and illustrate the motivation behind spherical masking in \Cref{sec:spherical_masking}. Furthermore, we evaluate the impact of different amounts of masked voxels in \Cref{sec:amount_voxel_mask} and discuss potential limitations in \Cref{sec:limitations}. 
Finally, we discuss additional insights on reconstruction results and data efficiency in \Cref{sec:additional_insights}. 

\section{Additional Results}
We provide results for CenterPoint~\cite{yinCenterBased3DObject2021} and PV-RCNN~\cite{shiPVRCNNPointVoxelFeature2020} to illustrate that our pre-trained initialization significantly enhances these baseline models. 
Following insights from the main manuscript (Section 4.1), we observe in \Cref{tab:data_efficiency_cp_pv_rcnn} that our MAELi pre-training effectively improves detection performance in a low-data regime where only a limited number of annotated samples are available for fine-tuning. 
In \Cref{tab:waymo_comparison_ap}, we report AP scores for \gls{wod}~\cite{sunScalabilityPerceptionAutonomous2020}, extending the results from Table 2 in the main manuscript.
Additionally, in \Cref{tab:kitti_full_r11}, we present our findings on the KITTI 3D dataset using the $R_{11}$ metric, and make comparisons with ALSO~\cite{boulchALSOAutomotiveLidar2023} and Occupancy-MAE~\cite{minOccupancyMAE2023}.

\label{sec:additional_results}
\begin{table*}[ht!]
	\scriptsize
	\centering
	\begin{tabular}{cp{3cm}|cccccccccc}						
		\toprule			
		\multirow{3}{*}{\textbf{Fraction}} & \multirow{3}{*}{\textbf{Method}} &  \multicolumn{10}{c}{\textbf{3D AP/APH (LEVEL 2)}}        \\
		&  & \multicolumn{2}{c}{\textbf{Gain}} & \multicolumn{2}{c}{\textbf{Overall}} & \multicolumn{2}{c}{\textbf{Vehicle}} &  \multicolumn{2}{c}{\textbf{Pedestrian}} &  \multicolumn{2}{c}{\textbf{Cyclist}}              \\
		& & \textbf{AP} & \textbf{APH} & \textbf{AP} & \textbf{APH} & \textbf{AP} & \textbf{APH} & \textbf{AP} & \textbf{APH} & \textbf{AP} & \textbf{APH} \\
		\midrule
		\multirow{4}{*}{\shortstack{1\%\\(791 frames)}} & Centerpoint~\cite{yinCenterBased3DObject2021} & - & - & 39.64 & 36.50 &   41.01 & 40.32 &  40.01 & 32.63 &  37.90 & 36.55 \\
		& \textbf{+ MAELi} &                 \cellcolor{petrol!60}+9.29 & \cellcolor{petrol!60}+8.75 &  \cellcolor{petrol!60}48.93 & \cellcolor{petrol!60}45.25 & \cellcolor{petrol!60}49.99 & \cellcolor{petrol!60}49.24 &  \cellcolor{petrol!60}51.92 & \cellcolor{petrol!60}43.07 &  \cellcolor{petrol!60}44.89 & \cellcolor{petrol!60}43.43 \\
		\hhline{~*{11}{-}}
		& PV-RCNN~\cite{shiPVRCNNPointVoxelFeature2020} & - & - & 43.93 & 30.72 &            51.34 & 48.70 &  41.59 & 20.35 &  38.86 & 23.11 \\
		& \textbf{+ MAELi} & \cellcolor{petrol!60}+7.53 & \cellcolor{petrol!60}+4.89 & \cellcolor{petrol!60}51.46 & \cellcolor{petrol!60}35.61 & \cellcolor{petrol!60}56.24 & \cellcolor{petrol!60}55.38 & \cellcolor{petrol!60}49.41 & \cellcolor{petrol!60}25.32 & \cellcolor{petrol!60}48.73 & \cellcolor{petrol!60}26.14 \\
		\hhline{*{12}{-}}`
		\multirow{4}{*}{\shortstack{5\%\\(3952 frames)}} & Centerpoint~\cite{yinCenterBased3DObject2021} &              - & - &  53.91 & 51.16 &          53.04 & 52.45 &  52.73 & 46.51 &  55.96 & 54.53 \\
		& \textbf{+ MAELi} &           \cellcolor{petrol!60}+4.49 & \cellcolor{petrol!60}+4.21 & \cellcolor{petrol!60}58.40 & \cellcolor{petrol!60}55.37 & \cellcolor{petrol!60}57.62 & \cellcolor{petrol!60}57.01 & \cellcolor{petrol!60}59.01 & \cellcolor{petrol!60}51.83 & \cellcolor{petrol!60}58.57 & \cellcolor{petrol!60}57.27 \\
		\hhline{~*{11}{-}}
		& PV-RCNN~\cite{shiPVRCNNPointVoxelFeature2020} &             - & - &  56.98 & 38.98 &           61.66 & 60.86 &  53.28 & 27.15 &  56.00 & 28.92 \\
		& \textbf{+ MAELi} &            \cellcolor{petrol!60}+1.64 & \cellcolor{petrol!60}+1.39 & \cellcolor{petrol!60}58.62 & \cellcolor{petrol!60}40.37 & \cellcolor{petrol!60}62.77 & \cellcolor{petrol!60}62.04 &  \cellcolor{petrol!60}57.07 & \cellcolor{petrol!60}29.05 & \cellcolor{petrol!60}56.02 & \cellcolor{petrol!60}30.02 \\
		\hhline{*{12}{-}}
		\multirow{4}{*}{\shortstack{10\%\\(7904 frames)}} & Centerpoint~\cite{yinCenterBased3DObject2021} &           - & - &   58.09 & 55.41 &            56.95 & 56.40 &  56.97 & 50.90 &  60.35 & 58.94 \\
		& \textbf{+ MAELi} &           \cellcolor{petrol!60}+3.26 & \cellcolor{petrol!60}+3.07 & \cellcolor{petrol!60}61.35 & \cellcolor{petrol!60}58.48 & \cellcolor{petrol!60}59.93 & \cellcolor{petrol!60}59.36 & \cellcolor{petrol!60}62.06 & \cellcolor{petrol!60}55.30 & \cellcolor{petrol!60}62.06 & \cellcolor{petrol!60}60.78 \\
		\hhline{~*{11}{-}}
		& PV-RCNN~\cite{shiPVRCNNPointVoxelFeature2020} &           - & - &   60.09 & \cellcolor{petrol!60}41.89 &            63.73 & 63.05 &  57.32 & 30.09 &  59.23 & \cellcolor{petrol!60}32.53 \\
		& \textbf{+ MAELi} & \cellcolor{petrol!60}+1.19 & -0.05 &  \cellcolor{petrol!60}61.28 & 41.84 &  \cellcolor{petrol!60}64.63 & \cellcolor{petrol!60}63.99 &  \cellcolor{petrol!60}59.82 & \cellcolor{petrol!60}30.90 &  \cellcolor{petrol!60}59.40 & 30.62 \\
		\hhline{*{12}{-}}
		\multirow{4}{*}{\shortstack{20\%\\(15808 frames)}} & Centerpoint~\cite{yinCenterBased3DObject2021} &            - & - &   61.81 & 59.15 &           60.59 & 60.06 &  61.19 & 55.03 &  63.64 & 62.36 \\
		& \textbf{+ MAELi} &            \cellcolor{petrol!60}+0.98 & \cellcolor{petrol!60}+0.90 & \cellcolor{petrol!60}62.79 & \cellcolor{petrol!60}60.05 & \cellcolor{petrol!60}61.79 & \cellcolor{petrol!60}61.23 & \cellcolor{petrol!60}63.47 & \cellcolor{petrol!60}57.04 & \cellcolor{petrol!60}63.11 & \cellcolor{petrol!60}61.87 \\
		\hhline{~*{11}{-}}
		& PV-RCNN~\cite{shiPVRCNNPointVoxelFeature2020} &            - & - &  62.15 & 42.99 &            65.01 & 64.35 &  60.40 & 30.30 &  61.05 & 34.32 \\
		& \textbf{+ MAELi} &            \cellcolor{petrol!60}+0.49 & \cellcolor{petrol!60}+7.21 &  \cellcolor{petrol!60}62.65 & \cellcolor{petrol!60}50.20 & \cellcolor{petrol!60}65.45 & \cellcolor{petrol!60}64.85 & \cellcolor{petrol!60}61.54 & \cellcolor{petrol!60}35.52 & \cellcolor{petrol!60}60.95 & \cellcolor{petrol!60}50.24 \\
		\bottomrule			
	\end{tabular}\\[2mm]	
	\caption{
	Quantitative results of our pre-training on Centerpoint and PV-RCNN on the Waymo \emph{val} set.
	For each detector, we report the results of training from scratch (upper row) and the improved results utilizing a MAELi-pre-trained initialization (lower row), respectively.
	We use the first 399 sequences of the Waymo \emph{train} set for pre-training and different fractions of the second 399 sequences for fine-tuning.}
	\label{tab:data_efficiency_cp_pv_rcnn}
\end{table*}

\begin{table*}
	\scriptsize
	\centering
	\begin{tabular}{p{1.7cm}|cccccccccc}
		\toprule
		\multirow{2}{*}{\textbf{Method}} &  \multicolumn{10}{c}{\textbf{3D AP/APH (LEVEL 2)}}        \\
		&  \multicolumn{2}{c}{\textbf{Gain}}  & \multicolumn{2}{c}{\textbf{Overall}} & \multicolumn{2}{c}{\textbf{Vehicle}} & \multicolumn{2}{c}{\textbf{Pedestrian}} & \multicolumn{2}{c}{\textbf{Cyclist}} \\
 		& \textbf{AP} & \textbf{APH} & \textbf{AP} & \textbf{APH} & \textbf{AP} & \textbf{APH} & \textbf{AP} & \textbf{APH} & \textbf{AP} & \textbf{APH} \\
		\midrule
		SECOND~\cite{yanSECONDSparselyEmbedded2018} & - & - &  58.26 & 54.35 &  62.58 & 62.02 &  57.22 & 47.49 &  54.97 & 53.53 \\
		+ Occ-MAE~\cite{minOccupancyMAE2023} & \cellcolor{petrol!25}+0.85 & \cellcolor{petrol!25}+0.75            &  \cellcolor{petrol!25}59.11 & \cellcolor{petrol!25}55.10 &  \cellcolor{petrol!25}62.67 & \cellcolor{petrol!25}62.34 &  \cellcolor{petrol!25}59.03 & \cellcolor{petrol!25}48.79 &  \cellcolor{petrol!25}55.62 & \cellcolor{petrol!25}54.17 \\
		\textbf{+ MAELi}  & \cellcolor{petrol!60}+2.32 & \cellcolor{petrol!60}+2.35 &  \cellcolor{petrol!60}60.57 & \cellcolor{petrol!60}56.69 &  \cellcolor{petrol!60}63.75 & \cellcolor{petrol!60}63.20 &  \cellcolor{petrol!60}60.71 & \cellcolor{petrol!60}50.93 &  \cellcolor{petrol!60}57.26 & \cellcolor{petrol!60}55.95 \\
		\hhline{*{11}{-}}
		CenterPoint~\cite{yinCenterBased3DObject2021}  & - & -             &  64.51 & 61.92 &  63.16 & 62.65 &  64.27 & 58.23 &  66.11 & 64.87 \\
		+ Occ-MAE~\cite{minOccupancyMAE2023} & \cellcolor{petrol!60}+1.35 & \cellcolor{petrol!60}+1.31 &  \cellcolor{petrol!60}65.86 & \cellcolor{petrol!60}63.23 &  \cellcolor{petrol!25}64.05 & \cellcolor{petrol!25}63.53 &  \cellcolor{petrol!25}65.78 & \cellcolor{petrol!25}59.62 &  \cellcolor{petrol!60}67.76 & \cellcolor{petrol!60}66.53 \\
		\textbf{+ MAELi} & \cellcolor{petrol!25}+1.09 & \cellcolor{petrol!25}+1.08     & \cellcolor{petrol!25}65.60 & \cellcolor{petrol!25}63.00 & \cellcolor{petrol!60}64.22 & \cellcolor{petrol!60}63.70 & \cellcolor{petrol!60}65.93 & \cellcolor{petrol!60}59.79 & \cellcolor{petrol!25}66.66 & \cellcolor{petrol!25}65.52 \\
		\hhline{*{11}{-}}
		PV-RCNN~\cite{shiPVRCNNPointVoxelFeature2020}  & - & - & 59.84 & 56.23 & 64.99 & 64.38 & 53.80 & 45.14 & 60.72 & 59.18 \\
		+ GCC-3D~\cite{liangExploringGeometryAwareContrast2021} & +1.46 & +1.95 & 61.30 & 58.18 & 65.65 & 65.10 & 55.54 & 48.02 & 62.72 & 61.43 \\
		+ PropCont~\cite{yinProposalContrastUnsupervisedPretraining2022} & +2.78 & +3.05 & 62.62 & 59.28 & 66.04 & 65.47 & 57.58 & 49.51 & \cellcolor{petrol!25}64.23 & \cellcolor{petrol!25}62.86 \\
		+ Occ-MAE~\cite{minOccupancyMAE2023} & \cellcolor{petrol!60}+5.99 & \cellcolor{petrol!25}+5.74 & \cellcolor{petrol!60}65.82 & \cellcolor{petrol!25}61.98 & \cellcolor{petrol!60}67.94 & \cellcolor{petrol!60}67.34 & \cellcolor{petrol!25}64.91 & \cellcolor{petrol!25}55.57 & \cellcolor{petrol!60}64.62 & \cellcolor{petrol!60}63.02 \\
		\textbf{+ MAELi} & \cellcolor{petrol!25}+5.88 & \cellcolor{petrol!60}+5.92 & \cellcolor{petrol!25}65.72 & \cellcolor{petrol!60}62.15 & \cellcolor{petrol!25}67.90 & \cellcolor{petrol!60}67.34 & \cellcolor{petrol!60}65.14 & \cellcolor{petrol!60}56.32 & 64.13 & 62.79 \\
		\bottomrule
	\end{tabular}\\	
	\caption{Performance comparison on the Waymo \emph{val} set trained on 20\% of the Waymo \emph{train} set including AP scores. We compare different detectors trained from scratch with their pendants utilizing pre-trained weights from GCC-3D, ProposalContrast, Occupancy-MAE and the proposed MAELi.
	}
	\label{tab:waymo_comparison_ap}
\end{table*}

\begin{table}
	\scriptsize
	\centering
	\begin{tabular}{p{1.7cm}|cccccccccc}						
		\toprule
		\textbf{Method} & \textbf{Pre-train} & \textbf{mAP} & \textbf{Car} & \textbf{Pedestrian} & \textbf{Cyclist} \\
		\midrule
		SECOND~\cite{yanSECONDSparselyEmbedded2018} & - & 66.25 & 78.62 & 52.98 & 67.15 \\
		+ Occ-MAE~\cite{minOccupancyMAE2023} & KITTI 3D & 66.71 & \cellcolor{petrol!60}78.90 & 53.14 & 68.08 \\
		+ ALSO~\cite{boulchALSOAutomotiveLidar2023} & nuScenes & 67.29 & 78.65 & 55.17 & 68.05 \\
		+ ALSO~\cite{boulchALSOAutomotiveLidar2023} & KITTI 3D & 66.86 & \cellcolor{petrol!25}78.78 & 53.57 & 68.22 \\
		+ ALSO~\cite{boulchALSOAutomotiveLidar2023} & KITTI360 & 67.40 & 78.63 & 54.23 & 69.35 \\
		\textbf{+ MAELi} & Waymo & \cellcolor{petrol!25}68.31 & 78.44 & \cellcolor{petrol!25}55.72 & \cellcolor{petrol!25}70.78 \\
		\textbf{+ MAELi} & KITTI 3D & 67.51 & 78.20 & 55.48 & 68.86 \\
		\textbf{+ MAELi} & KITTI360 & \cellcolor{petrol!60}68.74 & 78.44 & \cellcolor{petrol!60}56.00 & \cellcolor{petrol!60}71.79 \\
		\hhline{*{7}{-}}
		PV-RCNN~\cite{shiPVRCNNPointVoxelFeature2020} & - & 70.66 & 83.61 & 57.90 & 70.47 \\
		+ Occ-MAE~\cite{minOccupancyMAE2023} & KITTI 3D & 71.73 & \cellcolor{petrol!25}83.82 & 59.37 & 71.99 \\
		+ ALSO~\cite{boulchALSOAutomotiveLidar2023} & nuScenes & 72.20 & 83.77 & 58.49 & \cellcolor{petrol!60}74.35 \\
		+ ALSO~\cite{boulchALSOAutomotiveLidar2023} & KITTI 3D & 71.96 & 83.67 & 58.48 & 73.74 \\
		+ ALSO~\cite{boulchALSOAutomotiveLidar2023} & KITTI360 & \cellcolor{petrol!25}72.69 & 83.39 & \cellcolor{petrol!25}60.83 & \cellcolor{petrol!25}73.85 \\
		\textbf{+ MAELi} & Waymo & 71.79 & 83.38 & 58.53 & 73.45 \\
		\textbf{+ MAELi} & KITTI 3D & 70.70 & 79.22 & 60.02 & 72.87 \\
		\textbf{+ MAELi} & KITTI360 & \cellcolor{petrol!60}73.03 & \cellcolor{petrol!60}83.99 & \cellcolor{petrol!60}62.43 & 72.67 \\
		\bottomrule		
	\end{tabular}\\[2mm]	
	\caption{
		Quantitative results of our pre-training on SECOND and PV-RCNN on the KITTI 3D \emph{val} set using the $R_{11}$ metric.
	}
	\label{tab:kitti_full_r11}
\end{table}

\section{Sparse Reconstruction Decoder for 3D Object Detection}
\label{sec:architecture}

To describe the architecture of our decoder in detail, we group operations with the same \emph{voxel/tensor stride} into a \emph{block}.
In \Cref{tab:decoder_architecture}, we list the different decoder blocks in addition to the preceding \gls{bev} encoder (summarized as single entry) and the required \emph{reshaping+sampling} step to transform the dense feature representation back to a sparse 3D tensor.

\begin{table}[htb]
	\scriptsize
	\centering
	\begin{tabular}{l|ccc}
		\toprule
		\multirow{2}{*}{\textbf{Description}} & \multirow{2}{*}{\textbf{\# Channels}} & \textbf{Voxel/Tensor}  & \textbf{Spatial} \\		
		& & \textbf{Stride} &	 \textbf{Dimension} \\
		\midrule
		Output BEV Encoder & $512$ & - & $188 \times 188$ \\
		Reshaping + Sampling & $256$ & $8 \times 8 \times 16$ & $188 \times 188 \times 2$ \\
		DBlock 1 & $64$ & $8 \times 8 \times 8$ & $188 \times 188 \times 5$ \\
		DBlock 2 & $64$ & $4 \times 4 \times 4$ & $376 \times 376 \times 11$ \\
		DBlock 3 & $32$ & $2 \times 2 \times 2$ & $752 \times 752 \times 21$ \\
		DBlock 4 & $16$ & $1 \times 1 \times 1$ & $1504 \times 1504 \times 41$ \\
		\bottomrule
	\end{tabular}
	\caption{
		Architecture of our decoder. We state the number of channels, the voxel stride and the maximum spatial dimension for the Waymo Open Dataset \emph{after} each block.
		Stride and spatial dimensions are depicted in the format $x\times y \times z$.
		Each decoder block \emph{inverts} one downsampling step from the sparse 3D encoder, eventually resulting in the original voxel stride. 
	}
	\label{tab:decoder_architecture}
\end{table}

Each block comprises an upsampling step using \emph{generative transposed convolution} and a pruning step via $1 \times 1 \times 1$ \emph{submanifold sparse convolution}.
The operations are listed in \Cref{tab:decoder_block}.

\begin{table}[htb]
	\scriptsize
	\centering
	\begin{tabular}{l|cc}
		\toprule
		\textbf{Operation} & \textbf{Kernel Size} & \textbf{Stride}   \\			
		\midrule
		Generative Transposed Convolution & $2\times2\times2^\dagger$  &  $2\times2\times2^\dagger$ \\
		Batch Norm & - & - \\
		ReLU & - & - \\
		Submanifold Sparse Convolution & $3 \times 3 \times 3$  &  $1 \times 1 \times 1$ \\
		Batch Norm & - & - \\
		ReLU & - & - \\
		\midrule
		Submanifold Sparse Convolution & $1 \times 1 \times 1$  &  $1 \times 1 \times 1$ \\
		Pruning & - & - \\
		\bottomrule
	\end{tabular}
	\caption{
		Structure of each decoder block. We additionally state the operation's kernel size and stride, each in the format $x\times y \times z$. 	
		The upper part depicts the upsampling and feature transformation.
		The lower part uses the final feature representation from above and decides via classification whether a voxels is pruned or not.	
		$^\dagger$These values deviate for DBlock 1, where it has a kernel size of $1\times1\times3 $ and a stride of $1\times1\times2$ to invert the encoder's respective downsampling step. 
	}
	\label{tab:decoder_block}
\end{table}

\section{Spherical Masking - Illustration}
\label{sec:spherical_masking}
\begin{figure}[b]
	\centering
	\includegraphics[width=\linewidth]{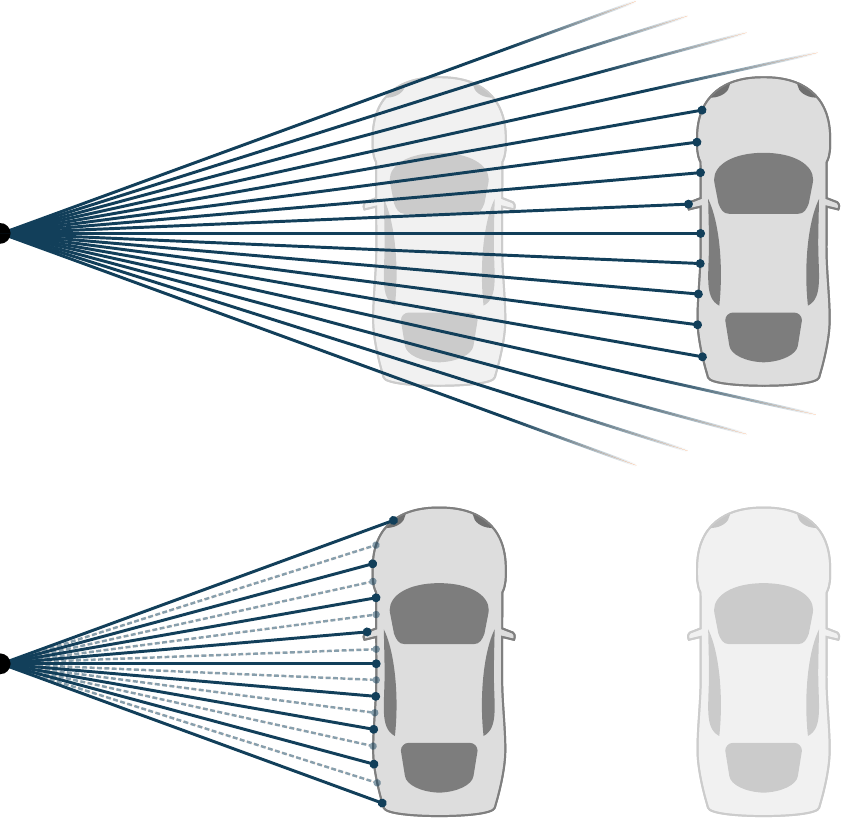}
	\caption{
		Spherical masking reduces the angular resolution of the LiDAR (bottom). The resulting sampling is thus similar to objects that are farther away (top).
    }
	\label{fig:spherical_masking}
\end{figure}

As discussed in the main manuscript (Section 3.3), spherical masking reduces the angular resolution in azimuth and inclination by subsampling the LiDAR's range image. \Cref{fig:spherical_masking} illustrates the effect of this sampling on the LiDAR's range image. 
We sample objects as if they were located at a larger distance. Since nearby objects are more densely sensed by the LiDAR, we have more knowledge about the actual occupancy and thus, can induce a stronger self-supervision signal. This helps to improve the model's ability to generalize to objects located farther away. 

\section{Ablation Study}
\label{sec:amount_voxel_mask}

\begin{table*}[b]
	\centering
	\begin{tabular}{c|cccccccc}
	\toprule
	\multirow{3}{*}{\textbf{Method}} & \multicolumn{8}{c}{\textbf{3D AP/APH (LEVEL 2)}}   \\
	&  \multicolumn{2}{c}{\textbf{Overall}} &  \multicolumn{2}{c}{\textbf{[0m, 30m)}} &  \multicolumn{2}{c}{\textbf{[30m, 50m)}} &  \multicolumn{2}{c}{\textbf{[50m, +inf)}} \\
	& \textbf{AP} & \textbf{APH} & \textbf{AP} & \textbf{APH} & \textbf{AP} & \textbf{APH} & \textbf{AP} & \textbf{APH} \\
	\midrule
	\textbf{MAELi} & \cellcolor{petrol!60}51.05 & \cellcolor{petrol!60}50.11 & \cellcolor{petrol!60}80.22 & \cellcolor{petrol!60}79.37 & \cellcolor{petrol!60}48.86 & \cellcolor{petrol!60}47.64 & \cellcolor{petrol!25}21.09 & \cellcolor{petrol!25}19.98 \\
	w/o DW & \cellcolor{petrol!25}50.91 & \cellcolor{petrol!25}49.85 & \cellcolor{petrol!25}79.95 & \cellcolor{petrol!25}79.04 & \cellcolor{petrol!25}48.62 & \cellcolor{petrol!25}47.20 & \cellcolor{petrol!60}21.47 & \cellcolor{petrol!60}20.31 \\	
	w/o LAR & 50.05 & 48.87 & 79.33 & 78.20 & 47.77 & 46.27 & 20.66 & 19.56 \\
	w/o VM & 48.99 & 47.86 & 78.45 & 77.46 & 46.43 & 45.05 & 19.47 & 18.23 \\
	w/o SM & 49.90 & 48.85 & 79.50 & 78.54 & 46.99 & 45.63 & 20.53 & 19.37 \\
	\midrule
	Baseline & 41.64 & 40.02 & 72.59 & 70.79 & 37.09 & 34.86 & 13.14 & 11.96 \\
	\bottomrule
	\end{tabular}\\[2mm]	
	\caption{Impact of the components of MAELi evaluated on the Waymo \emph{val} set for \emph{Vehicle}. 
		We disable \emph{distance weighting} (w/o DW), \emph{LiDAR-aware reconstruction} (w/o LAR), \emph{voxel masking} (w/o VM) and \emph{spherical masking} (w/o SM).
		We use the first 399 sequences of the Waymo \emph{train} set for pre-training and 1\% of the second 399 sequences for fine-tuning.
		We utilize a SECOND~\cite{yanSECONDSparselyEmbedded2018} model and state a version trained from scratch as baseline.
	}
	\label{tab:ablation}
\end{table*}
\begin{table*}[htb]
	\centering
	\begin{tabular}{c|cccccccc}
		\toprule
		\multirow{3}{*}{\shortstack{\textbf{Fraction}\\\textbf{Voxels}}} & \multicolumn{8}{c}{\textbf{3D AP/APH (LEVEL 2)}}   \\
		& \multicolumn{2}{c}{\textbf{Overall}} & \multicolumn{2}{c}{\textbf{Vehicle}} & \multicolumn{2}{c}{\textbf{Pedestrian}} & \multicolumn{2}{c}{\textbf{Cyclist}} \\
		& \textbf{AP} & \textbf{APH} & \textbf{AP} & \textbf{APH} & \textbf{AP} & \textbf{APH} & \textbf{AP} & \textbf{APH} \\
		\midrule
		0.8 & 45.34 & \cellcolor{petrol!25}32.84 & 50.26 & 49.10 & 48.03 & 24.33 & 37.74 & \cellcolor{petrol!60}25.09 \\
		0.7 & 44.91 & 32.10 & 50.35 & 49.19 & 47.54 & 24.29 & 36.83 & 22.83 \\
		0.6 & \cellcolor{petrol!60}46.01 & \cellcolor{petrol!60}33.05 & \cellcolor{petrol!60}51.05 & \cellcolor{petrol!60}50.11 & \cellcolor{petrol!25}48.13 & \cellcolor{petrol!60}24.65 & \cellcolor{petrol!60}38.86 & \cellcolor{petrol!25}24.38 \\
		0.5 & 45.60 & 32.27 & \cellcolor{petrol!25}50.87 & \cellcolor{petrol!25}49.79 & 47.08 & 23.72 & \cellcolor{petrol!60}38.86 & 23.31 \\
		0.4 & \cellcolor{petrol!25} 45.79 & 31.85 & 50.36 & 49.24 & \cellcolor{petrol!60}48.27 & \cellcolor{petrol!25}24.50 & 38.74 & 21.81 \\
		\midrule
		Baseline &31.09& 22.25 & 41.64 & 40.02 & 33.39 & 17.45 & 18.24 & ~~9.29 \\
		\bottomrule
	\end{tabular}
	\caption{
		Impact of different amounts of voxels kept during \emph{voxel masking} evaluated on the Waymo \emph{val} set for \emph{Vehicle}.
		We use the first 399 sequences of the Waymo \emph{train} set for pre-training and 1\% of the second 399 sequences for fine-tuning.
		We utilize a SECOND~\cite{yanSECONDSparselyEmbedded2018} model and state a version trained from scratch as baseline.
	}
	\label{tab:ablation_voxel_crop}
\end{table*}

\noindent\textbf{Analyzing Pipeline Components:} We perform various experiments to investigate specific aspects of our pipeline, such as assessing the influence of our \emph{distance weighting} for empty voxels, evaluating our \emph{LiDAR-aware reconstruction} objective, and comparing the effectiveness of the \emph{voxel-} and \emph{spherical masking} strategies.

Therefore, we pre-train according to the \emph{Data Efficiency} protocol depicted in the main manuscript (Section 4.1) and fine-tune a SECOND~\cite{yanSECONDSparselyEmbedded2018} model on 1\% of the latter 399 sequences of the Waymo \emph{train} set.
To disable \emph{distance weighting}, we set $w_j^{D,\mathbf{s}} = 1$ for all \emph{empty} voxels $\mathbf{v}_{j}^{D,\mathbf{s}}$. 
To disable our \emph{LiDAR-aware reconstruction}, we additionally consider all \emph{unknown} voxels as \emph{empty}. 

We state the results in \Cref{tab:ablation} on \emph{Vehicle} LEVEL 2 across the distance ranges [0m, 30m), [30m, 50m) and [50m, $+\infty$).
Our \emph{LiDAR-aware reconstruction} objective improves the overall results across all ranges. 
While \emph{voxel masking} naturally has a nearly equal impact over all ranges, our \emph{spherical masking} is especially beneficial for the range [30m, 50m) with a gain of 1.87AP and 2.01APH.
The overall lower impact on the far distance range (above 50m) is also reasonable, since at this distance only very few points are sampled on the same object.

\noindent\textbf{Masking:} We evaluated our \emph{voxel masking} for different amounts of voxels. 
We maintain the training and evaluation scheme from above and vary the amount of kept voxels. 
The results are shown in \Cref{tab:ablation_voxel_crop}. 
Keeping $60\%$ of the voxels leads to the best overall results, eventually used for all other experiments with MAELi. 
However, in combination with \emph{spherical masking} significantly fewer points than this fraction actually remain.
To get an estimate, we evaluated the amount of effectively used points over $10000$ iterations, resulting in a fraction of $15.57\%$ on average.

\begin{figure*}[t]
	\centering
	\newcommand{\gsize}{0.49}
	\begin{subfigure}{\gsize\textwidth}
		\includegraphics[width=\linewidth]{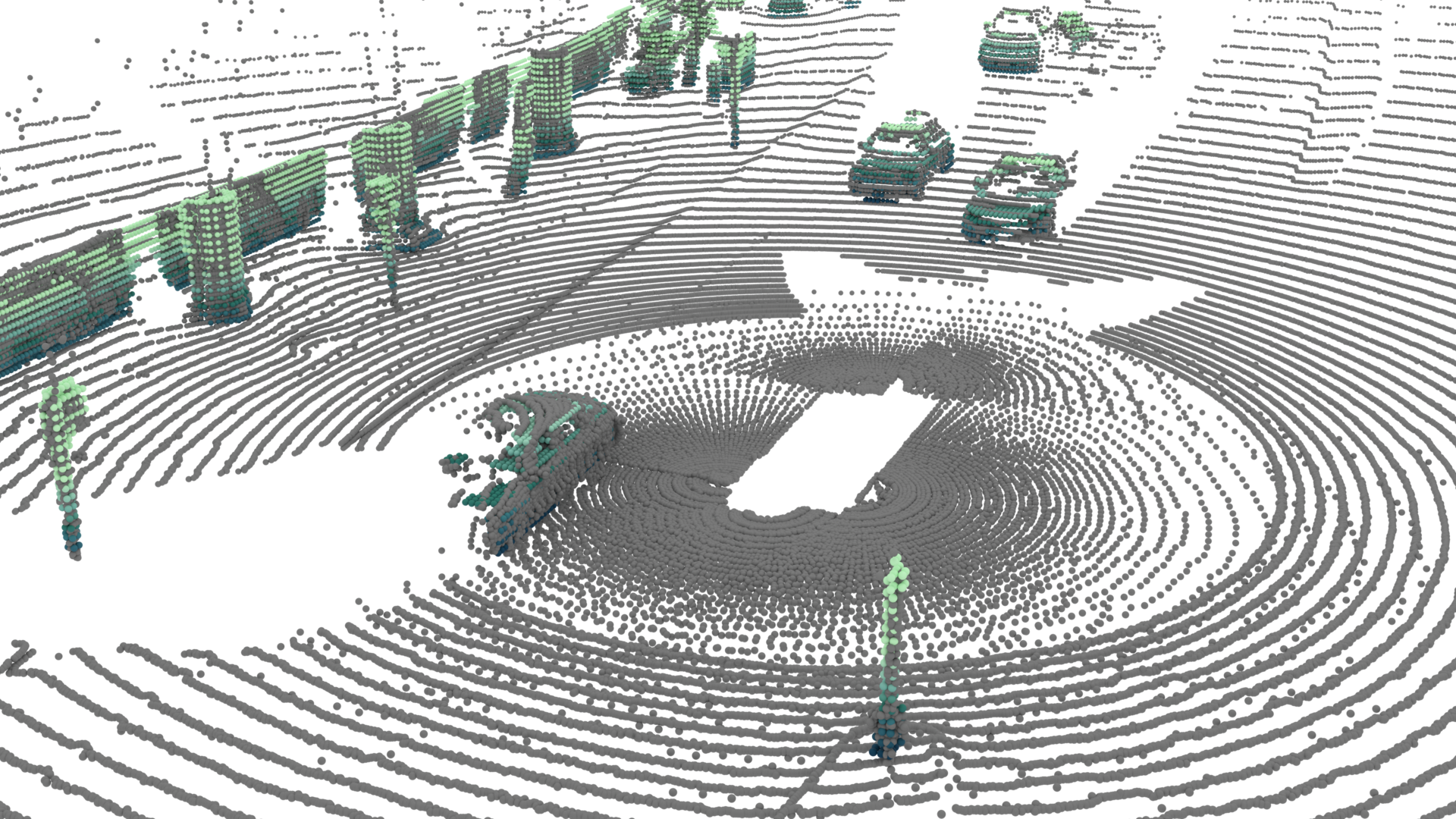}
		\caption{}
		\label{subfig:wo_lal}
	\end{subfigure}	
	\begin{subfigure}{\gsize\textwidth}
		\includegraphics[width=\linewidth]{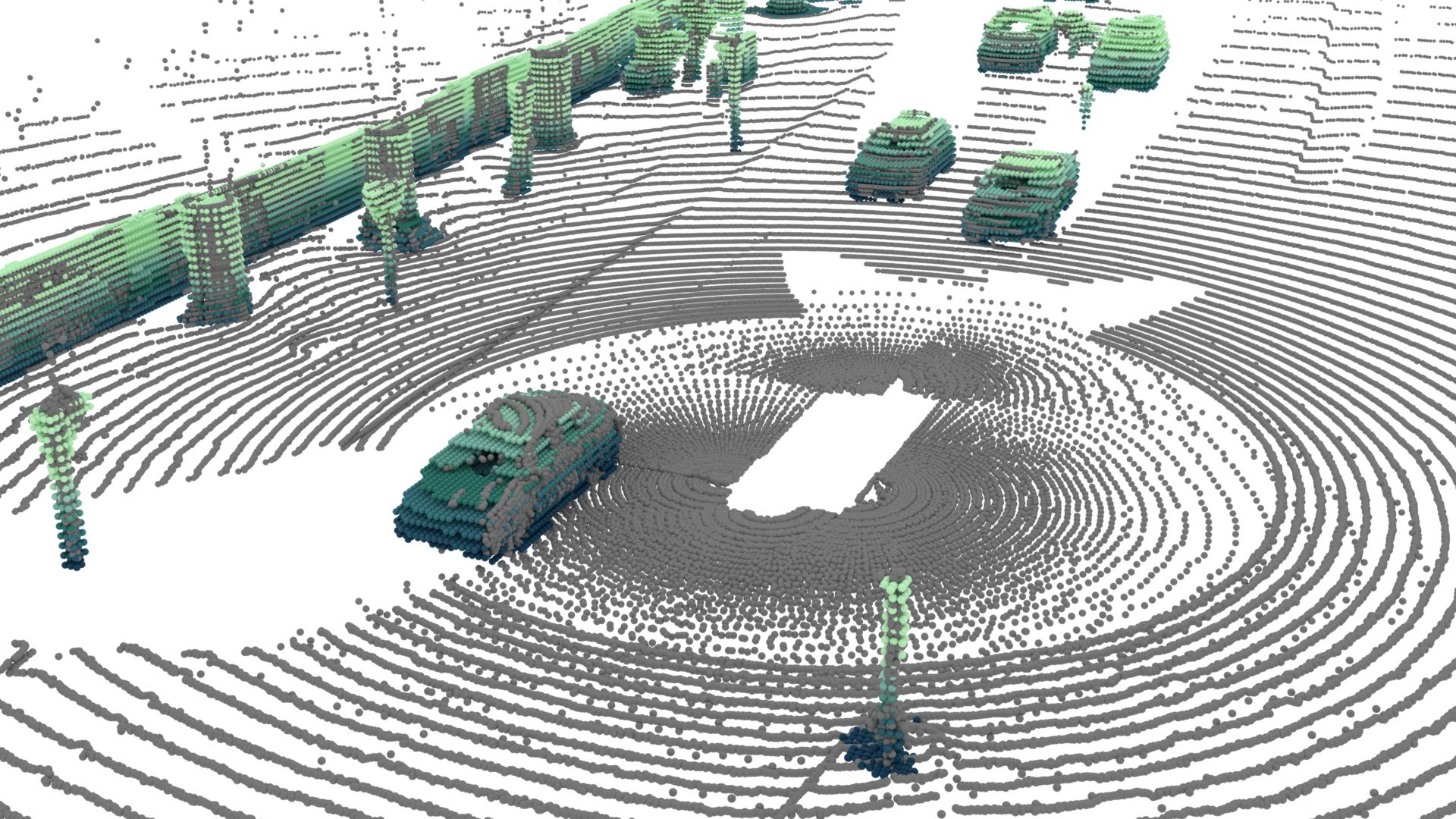}
		\caption{}
		\label{subfig:w_lal}
	\end{subfigure}	
	\caption{
		Completion results \subref{subfig:wo_lal} without and \subref{subfig:w_lal} with our \emph{LiDAR-aware reconstruction} on a full point cloud (gray). For visualization purposes, we color-coded the output points by their $z$-coordinate and removed the reconstructed ground plane.
	}
	\label{fig:completion}
\end{figure*}
\begin{figure*}[t]
	\centering
	\newcommand{\gsize}{0.49}
		\includegraphics[width=\gsize\linewidth,trim={5cm 0 5cm 0},clip]{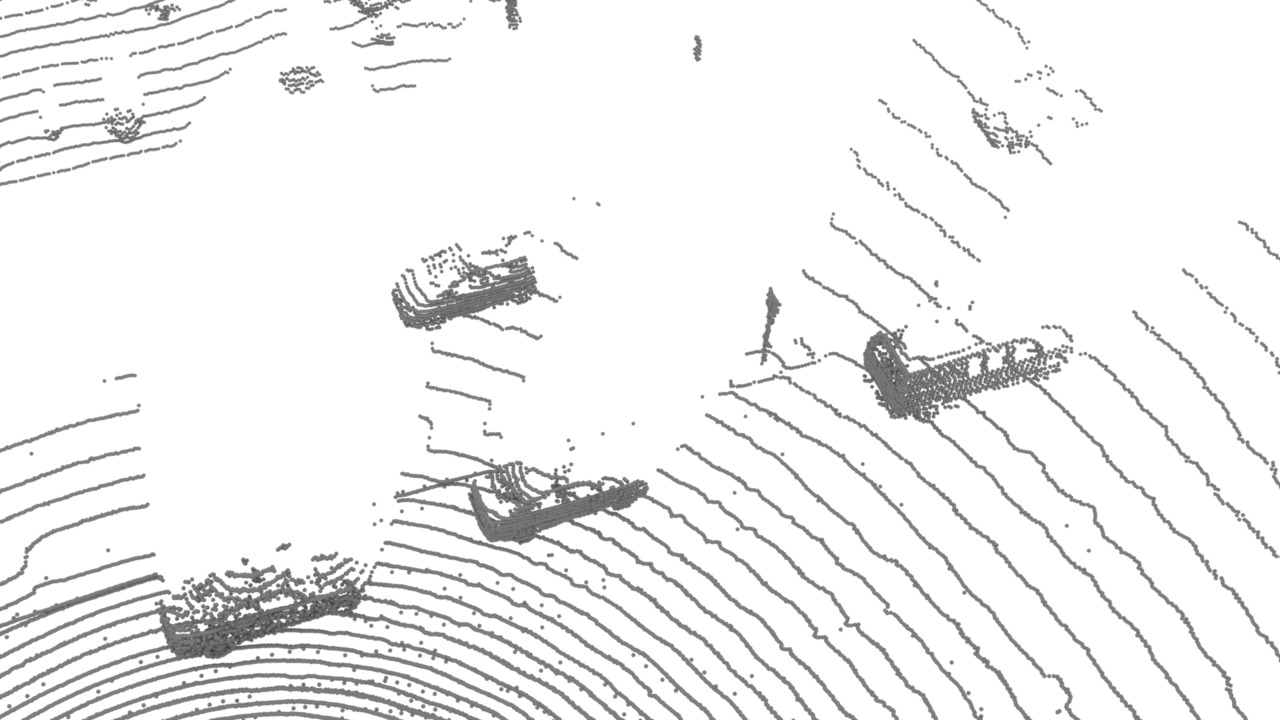}
		\includegraphics[width=\gsize\linewidth,trim={5cm 0 5cm 0},clip]{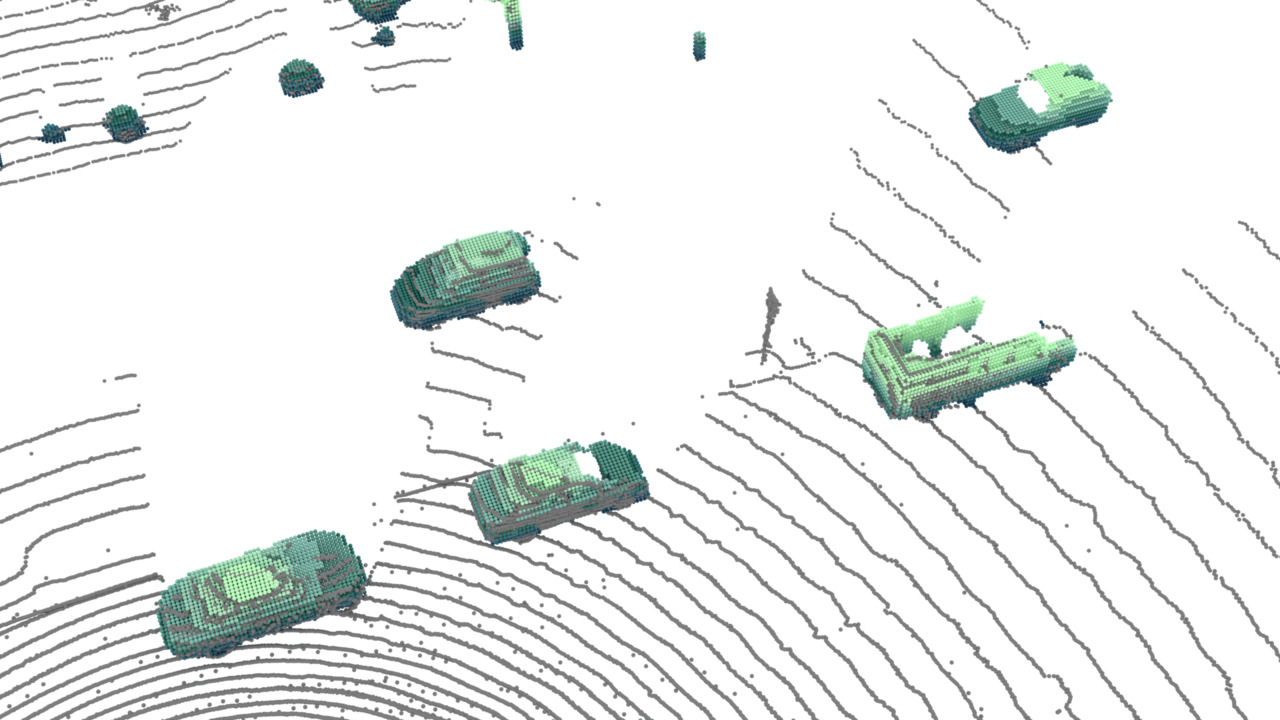}
		\includegraphics[width=\gsize\linewidth,trim={1cm 0 9cm 0},clip]{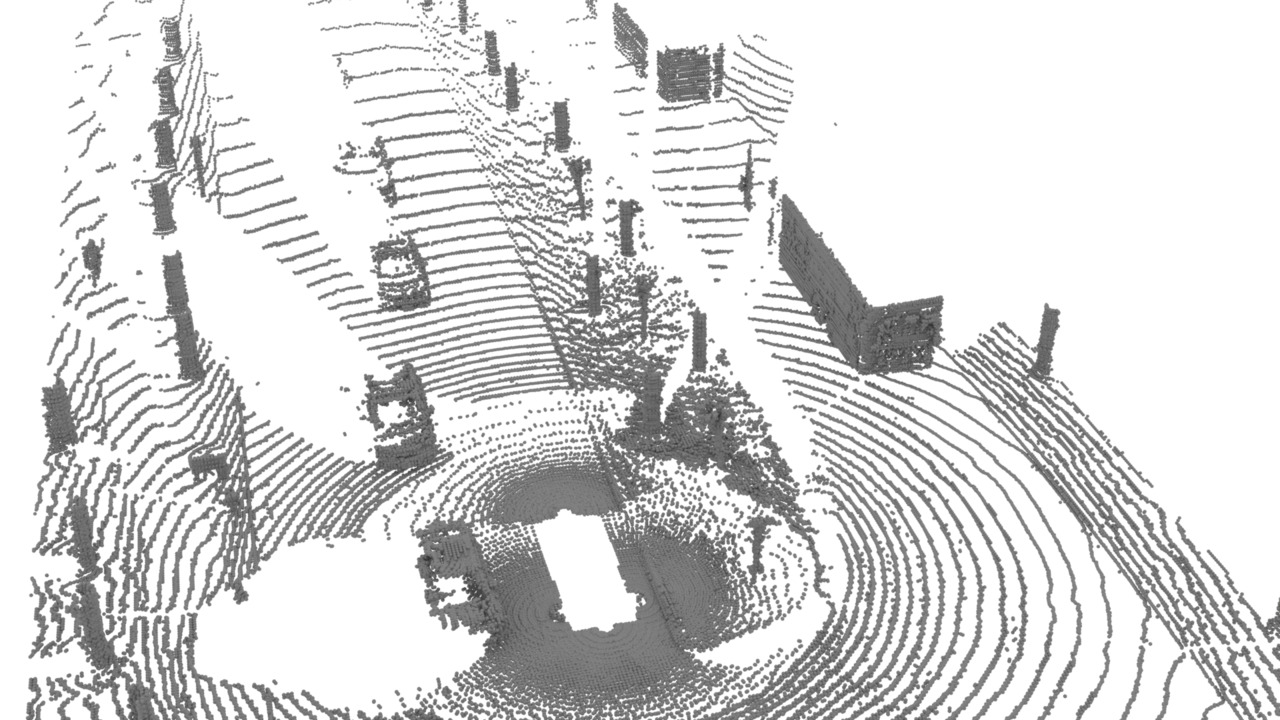}
		\includegraphics[width=\gsize\linewidth,trim={1cm 0 9cm 0},clip]{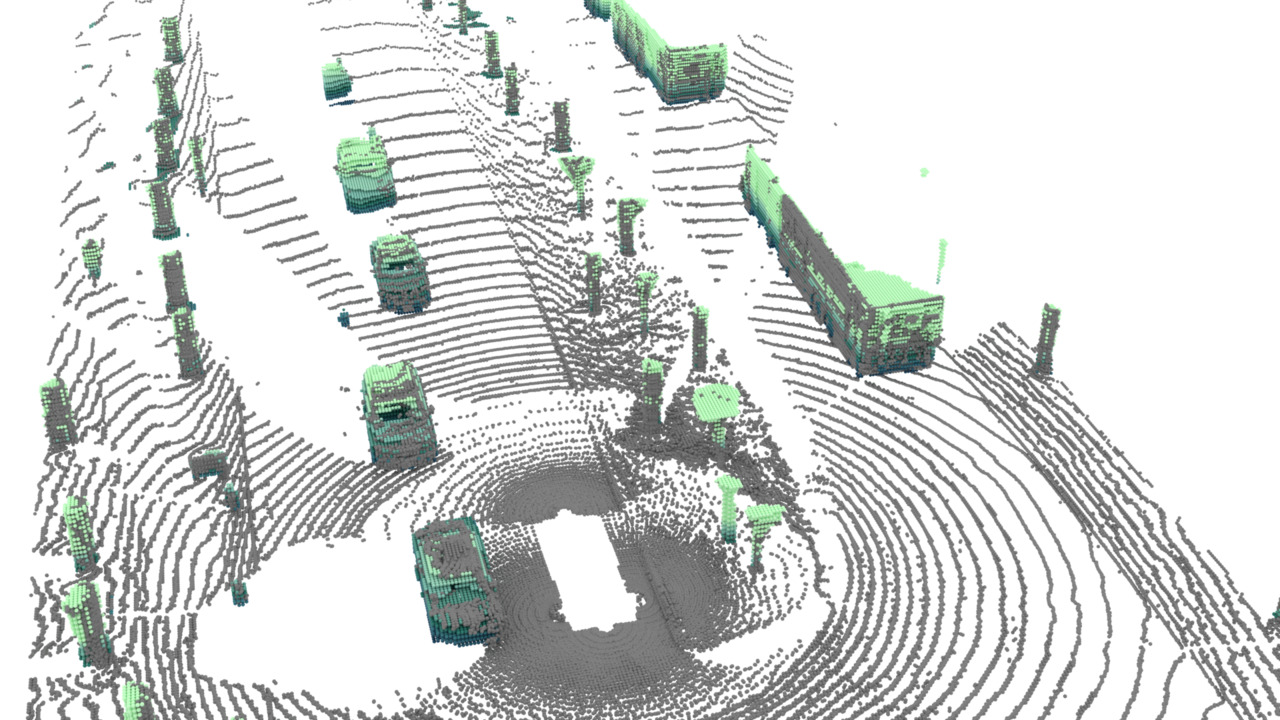}
		\includegraphics[width=\gsize\linewidth,trim={5cm 0 5cm 0},clip]{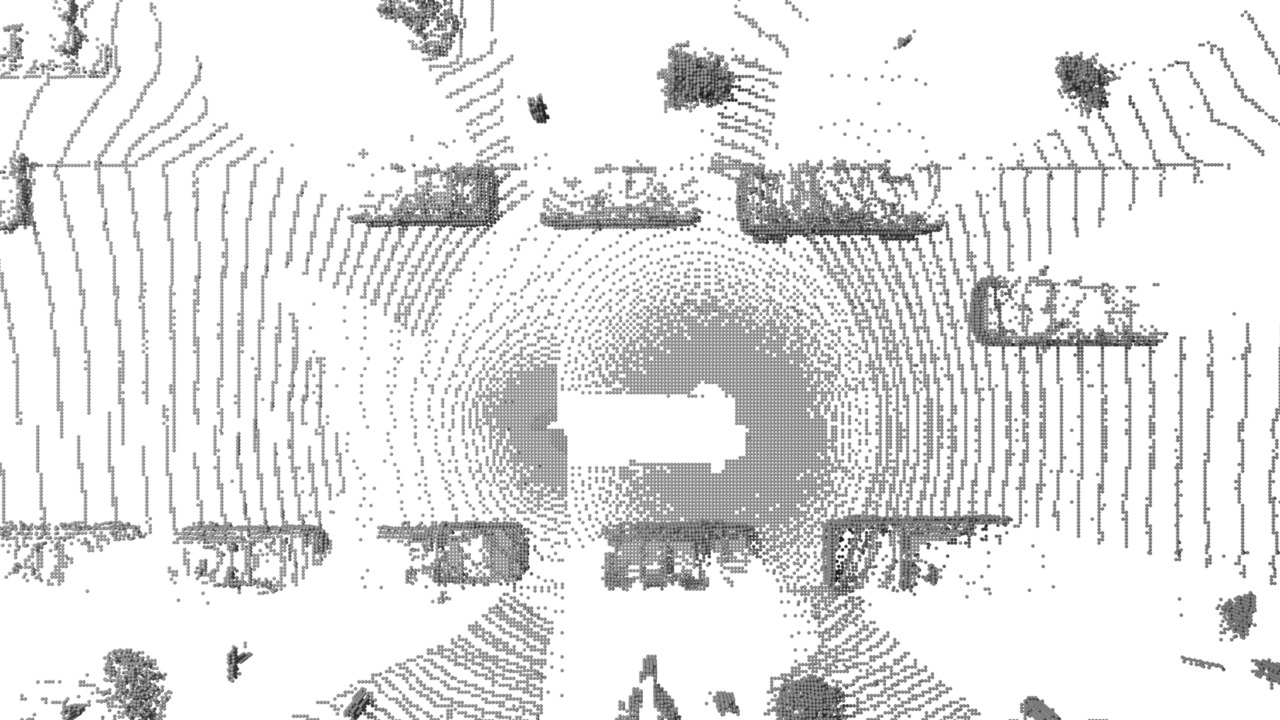}
		\includegraphics[width=\gsize\linewidth,trim={5cm 0 5cm 0},clip]{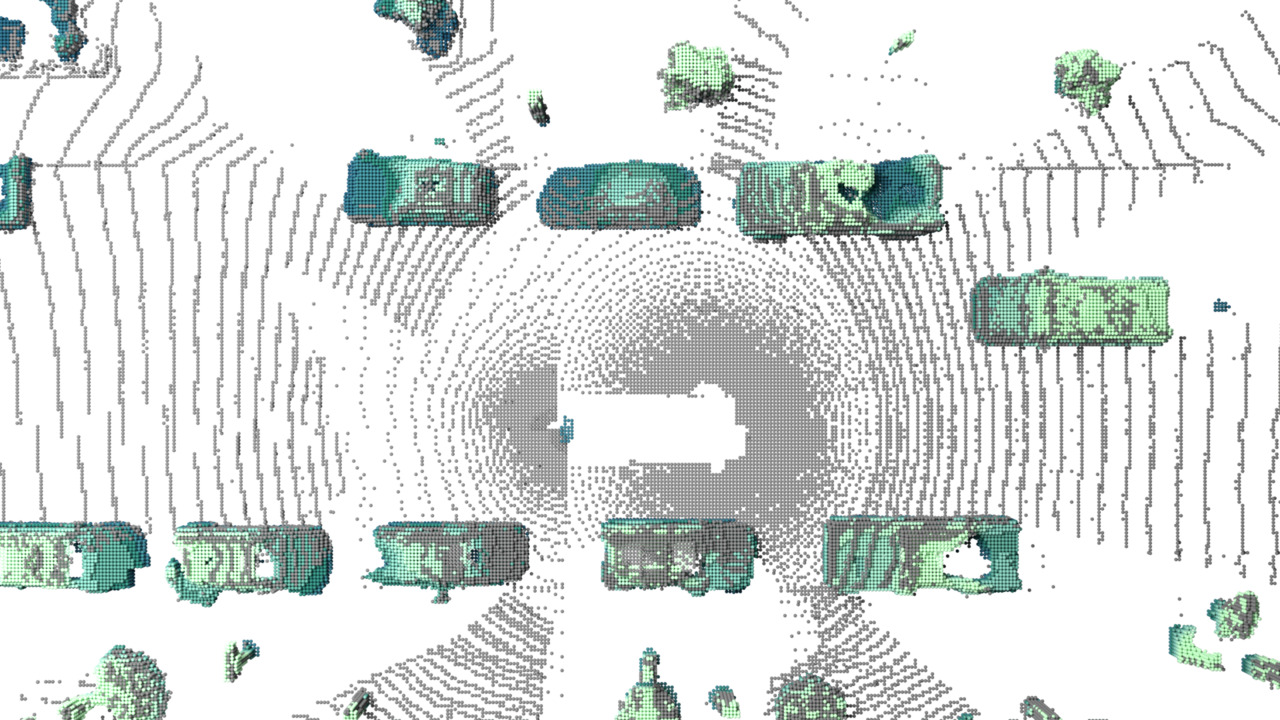}
	\caption{
		Further reconstruction results. We show the input point cloud on the left and the completed point cloud on the right. MAELi's reconstruction exhibits imperfections when reconstructing objects that are sparsely sampled or less frequent, such as trucks. However, it shows an apparent understanding of traffic scenes beyond a LiDAR's 2.5D sampling, \eg by symmetrically completing occluded parts of cars and poles. For visualization purposes, we color-coded the output points by their $z$-coordinate and removed the reconstructed ground plane.
	}
	\label{fig:completion_negatives}
\end{figure*}

\begin{figure*}[t]
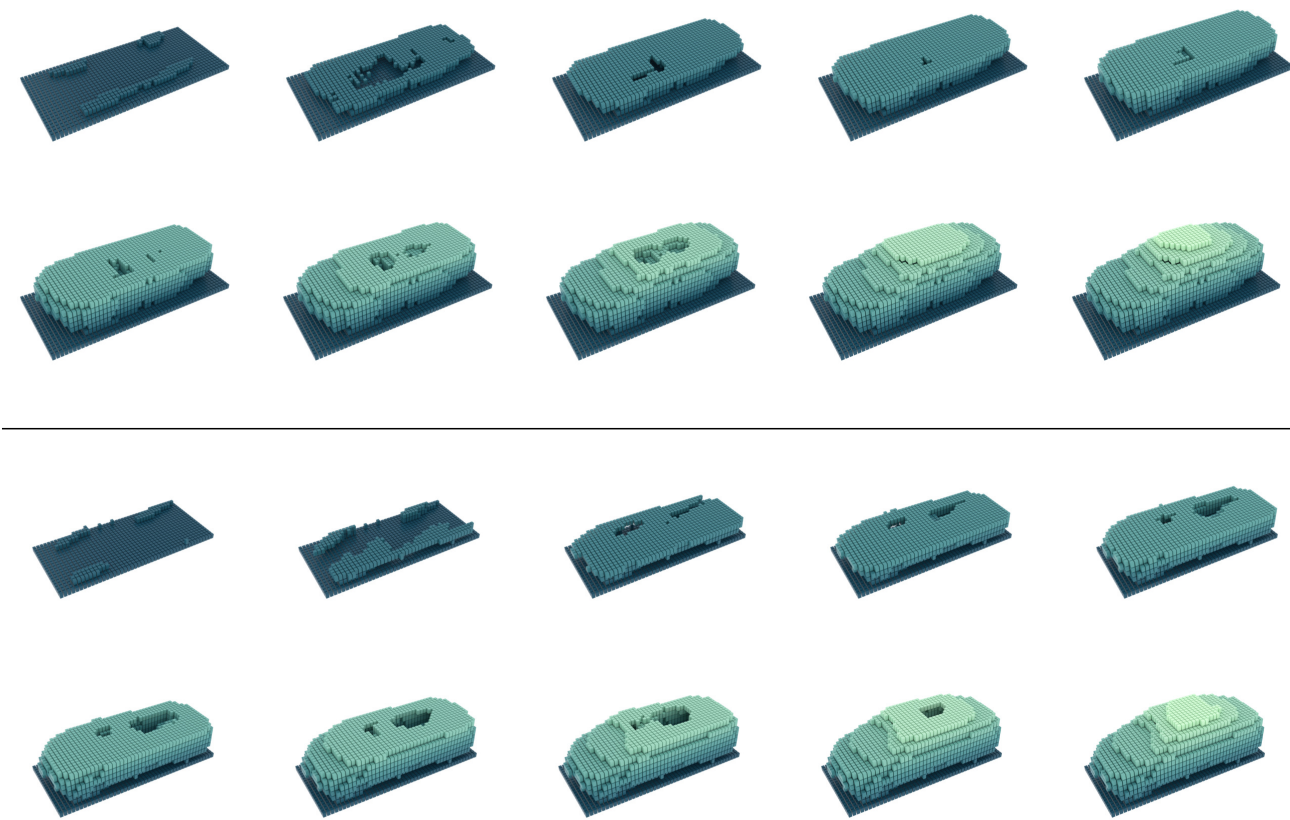

	\centering
	\newcommand{\gsize}{0.19}
	\newcounter{filename}	
	\begin{subfigure}{\textwidth}
		\forloop{filename}{3}{\value{filename} < 13}{
			\begin{subfigure}{\gsize\textwidth}
				\includegraphics[width=\linewidth]{gfx/reconstruction/supplementary/00\twodigits{\value{filename}}.jpg}
			\end{subfigure}
		}
	\end{subfigure}
	\rule{\textwidth}{0.2mm}
	\begin{subfigure}{\textwidth}
		\forloop{filename}{17}{\value{filename} < 27}{
			\begin{subfigure}{\gsize\textwidth}
				\includegraphics[width=\linewidth]{gfx/reconstruction/supplementary/00\twodigits{\value{filename}}.jpg}
			\end{subfigure}
		}
	\end{subfigure}
	\caption{
		Different layers of two reconstructed cars. 
		We observed that the reconstructed cars are often hollow. 
		There are visible tendencies to leave parts of the interior free.
	}
	\label{fig:border_reconstruction}
\end{figure*}

\pgfplotsset{width=7cm, height=6cm, compat=1.10}
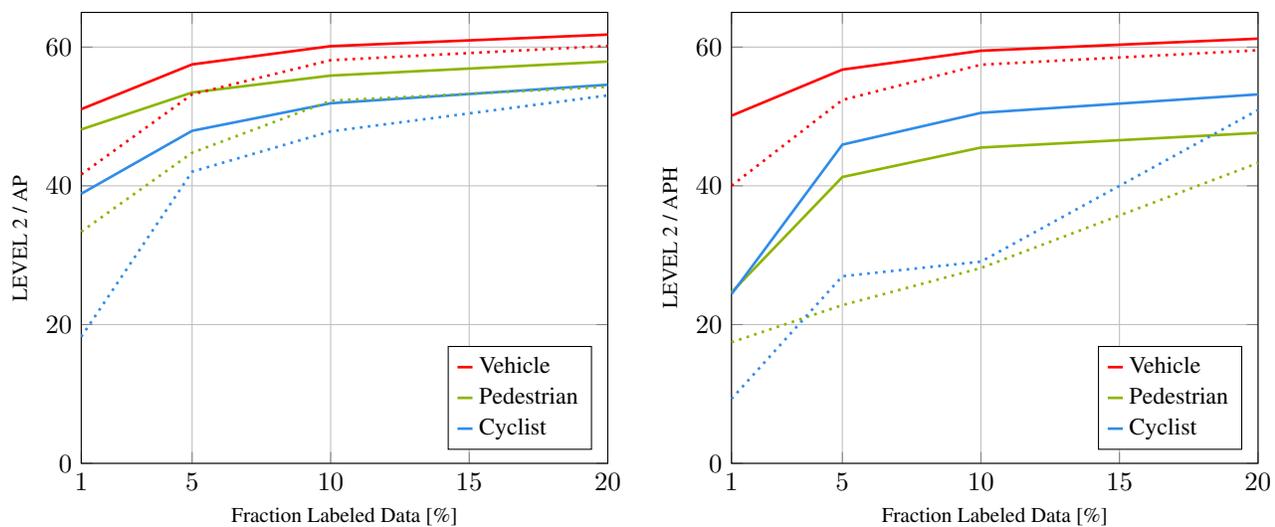
\begin{figure*}[tb]
	\centering
	\begin{subfigure}{0.49\textwidth}
		\begin{tikzpicture}
			\definecolor{bleudefrance}{rgb}{0.19, 0.55, 0.91}
			\definecolor{applegreen}{rgb}{0.55, 0.71, 0.0}
			\pgfplotsset{compat=1.11, set layers, ticks label/.append style ={blue},
				legend image code/.code={
					\draw[mark repeat=2,mark phase=2]
					plot coordinates {
						(0cm,0cm)
						(0.1cm,0cm)        %
						(0.2cm,0cm)         %
					};%
			}}
			\begin{axis}[
				legend cell align={left},
				legend pos=south east,
				grid,
				scale only axis,
				xmin=1,xmax=20,
				ymin=0, ymax=65,
				ylabel={\footnotesize LEVEL 2 / AP},
				xlabel={\footnotesize Fraction Labeled Data [\%]},
				xtick={1,5,10,15,20}]
				\addplot [name path=A, line width=1pt, red, mark options={scale=0.7}] table {gfx/training_progress/second_vehicle_l2ap_pret.txt};
				\addplot [name path=C, line width=1pt, applegreen, mark options={scale=0.7}] table {gfx/training_progress/second_pedestrian_l2ap_pret.txt};
				\addplot [name path=E, line width=1pt, bleudefrance, mark options={scale=0.7}] table {gfx/training_progress/second_cyclist_l2ap_pret.txt};
				\addplot [dotted, name path=B, line width=1pt, red, mark options={scale=0.7}] table {gfx/training_progress/second_vehicle_l2ap_vanilla.txt};
				\addplot [dotted, name path=D, line width=1pt, applegreen, mark options={scale=0.7}] table {gfx/training_progress/second_pedestrian_l2ap_vanilla.txt};
				\addplot [dotted, name path=F, line width=1pt, bleudefrance, mark options={scale=0.7}] table {gfx/training_progress/second_cyclist_l2ap_vanilla.txt};
				\addlegendentry{\small Vehicle}
				\addlegendentry{\small Pedestrian}
				\addlegendentry{\small Cyclist}
			\end{axis}
		\end{tikzpicture}
	\end{subfigure}
	\begin{subfigure}{0.49\textwidth}
		\begin{tikzpicture}
			\definecolor{bleudefrance}{rgb}{0.19, 0.55, 0.91}
			\definecolor{applegreen}{rgb}{0.55, 0.71, 0.0}
			\pgfplotsset{compat=1.11, set layers, ticks label/.append style ={blue},
				legend image code/.code={
					\draw[mark repeat=2,mark phase=2]
					plot coordinates {
						(0cm,0cm)
						(0.1cm,0cm)        %
						(0.2cm,0cm)         %
					};%
			}}
			\begin{axis}[
				legend cell align={left},
				legend pos=south east,
				grid,
				scale only axis,
				xmin=1,xmax=20,
				ymin=0, ymax=65,
				ylabel={\footnotesize LEVEL 2 / APH},
				xlabel={\footnotesize Fraction Labeled Data [\%]},
				xtick={1,5,10,15,20}]
				\addplot [name path=A, line width=1pt, red, mark options={scale=0.7}] table {gfx/training_progress/second_vehicle_l2aph_pret.txt};
				\addplot [name path=C, line width=1pt, applegreen, mark options={scale=0.7}] table {gfx/training_progress/second_pedestrian_l2aph_pret.txt};
				\addplot [name path=E, line width=1pt, bleudefrance, mark options={scale=0.7}] table {gfx/training_progress/second_cyclist_l2aph_pret.txt};
				\addplot [dotted, name path=B, line width=1pt, red, mark options={scale=0.7}] table {gfx/training_progress/second_vehicle_l2aph_vanilla.txt};
				\addplot [dotted, name path=D, line width=1pt, applegreen, mark options={scale=0.7}] table {gfx/training_progress/second_pedestrian_l2aph_vanilla.txt};
				\addplot [dotted, name path=F, line width=1pt, bleudefrance, mark options={scale=0.7}] table {gfx/training_progress/second_cyclist_l2aph_vanilla.txt};
				\addlegendentry{\small Vehicle}
				\addlegendentry{\small Pedestrian}
				\addlegendentry{\small Cyclist}
			\end{axis}
		\end{tikzpicture}
	\end{subfigure}
	\caption{
		Results of our pre-training on SECOND~\cite{yanSECONDSparselyEmbedded2018} on the Waymo \emph{val} set, using different amounts of labeled data. 
		We use the first 399 sequences of the Waymo \emph{train} set without any labels for pre-training and different fractions of the latter 399 sequences for fine-tuning. 
		The \emph{solid lines} are the results utilizing our pre-training with MAELi. The \emph{dotted lines} denote the version trained from scratch. 
	}
	\label{fig:data_efficiency}
\end{figure*}

\section{Limitations}
\label{sec:limitations}

Even though our sparse decoder allows for a memory efficient reconstruction, the amount of reconstructed voxels is obviously constrained by the available compute infrastructure.
Especially during the first iterations of our pre-training, while not sufficiently trained, some samples may lead to an uncontrolled reconstruction.
In order to regulate the amount of reconstructed voxels and to avoid training breakdowns, we introduce two limiting factors.
First, we estimate an average ground plane for each dataset and prune all reconstructed voxels that are $0.1m$ below this plane, as these generally do not contribute valuable information.
Second, we introduce a threshold for the maximum amount of reconstructed voxels to ensure that we do not run into memory issues.
If an upsampling step would generate more voxels than this limit, we randomly prune before the upsampling.
For these pruned voxels, simply no loss is induced, which only slightly \emph{delays} the training effect for these rare cases. 
For the detection experiments, we set the maximum number of total voxels to 6 million, which are easily processable, \eg on an NVIDIA\textsuperscript{\textregistered} GeForce\textsuperscript{\textregistered} RTX 3090 GPU.
We counted only 62 limit exceedances within the first 10k iterations of a random experiment. 

\section{Additional Insights}
\label{sec:additional_insights}

\noindent\textbf{Reconstruction Capabilities:} In \Cref{fig:completion}, we visualize the reconstruction capabilities of our \emph{LiDAR-aware loss} on a full point cloud. 
It encourages the network to fill up gaps in the wall and reconstruct the occluded areas of cars.

\Cref{fig:completion_negatives} highlights that reconstruction outcomes can vary across different objects, with some objects such as the less frequent trucks presenting greater challenges. However, the MAELi model demonstrates an advanced understanding of object semantics, enabling it to complete objects beyond the visible LiDAR input point cloud. 

In \Cref{fig:border_reconstruction}, we show the individual layers of two different reconstructed cars. 
We can see that our pre-training approach indeed encourages the model to go beyond the sampled LiDAR surface, reconstructing the entire car, also showing some hints for the correct placement of tires. 
Furthermore, especially parts of the interior, which are often surrounded by glass and thus, sometimes traversed by LiDAR beams, are seemingly left out during the reconstruction.
\\\\
\noindent\textbf{AP vs APH Data Efficiency:} In \Cref{fig:data_efficiency}, we plot the data efficiency results on Waymo~\cite{sunScalabilityPerceptionAutonomous2020} using our MAELi-based pre-training on the SECOND~\cite{yanSECONDSparselyEmbedded2018} detector. 
All three classes benefit from our pre-training (solid lines) compared to the vanilla version trained from scratch (dotted lines). 
With little data, the vanilla detector especially struggles to estimate the heading, which can be clearly seen for the smaller, less represented classes \emph{Pedestrian} and \emph{Cyclist}. 
However, utilizing our initialization, a proper estimation and significant detection improvements are possible already, early on, with only few annotated data samples.

\end{document}